\documentclass[11pt]{article}

\usepackage[letterpaper,margin=1in,headheight=14pt]{geometry}
\usepackage[T1]{fontenc}
\usepackage[utf8]{inputenc}
\usepackage{lmodern}
\usepackage{microtype}
\usepackage{etoolbox}
\usepackage{amsmath,amssymb}
\usepackage{array}
\usepackage{ragged2e}
\usepackage{booktabs}
\usepackage{needspace}
\usepackage{calc}
\usepackage{longtable}
\usepackage{caption}
\usepackage{graphicx}
\usepackage{listings}
\usepackage{xcolor}
\usepackage{xurl}
\usepackage{mslapa}
\usepackage[
  colorlinks=true,
  linkcolor=blue!45!black,
  citecolor=blue!45!black,
  urlcolor=blue!45!black,
  pdfborder={0 0 0}
]{hyperref}

\providecommand{\passthrough}[1]{#1}

\providecommand{\citewithlabel}[2]{\cite(#1;){#2}}
\providecommand{\shortcitewithlabel}[2]{\shortcite(#1;){#2}}

\hypersetup{
  pdftitle={What Reaches Expert Review? Representation, Structural Screening, and Candidate-Form Dependence in AI-Assisted Item Development},
  pdfauthor={Christopher Brooks},
  pdfsubject={Computational psychometrics and AI-assisted item development},
  pdfkeywords={AI-assisted item generation; computational psychometrics; text embeddings; item reduction; content representation}
}

\IfFileExists{glyphtounicode.tex}{%
  \input{glyphtounicode}
  \pdfgentounicode=1
}{}

\renewcommand{\arraystretch}{1.08}
\renewcommand{\raggedright}{\RaggedRight\hspace{0pt}}
\AtBeginEnvironment{longtable}{\scriptsize}

\title{
What Reaches Expert Review? Representation, Structural Screening, and Candidate-Form Dependence in AI-Assisted Item Development
}
\author{%
  Christopher Brooks\\
  \small School of Information, University of Michigan\\
  \small \href{mailto:brooksch@umich.edu}{brooksch@umich.edu}\quad\textbar\quad
  \href{https://orcid.org/0000-0003-0875-0204}{ORCID 0000-0003-0875-0204}
}
\date{August 2026}

\begin{document}

\maketitle

\begin{abstract}
Between AI-assisted item generation and expert review sits a computational evaluator whose decisions are usually treated as technical preliminaries. Yet representation, structural reduction, and selection policy determine which items and evidence psychometricians ever receive. Across two linked in-silico studies of 32,000 selected Big Five items, we followed fixed source populations from semantic representation through structural evaluation and candidate-form construction. Broad agreement in semantic geometry concealed consequential local differences: identical wording acquired different construct evidence, different items survived, and intended attributes could disappear even as community correspondence improved. These sensitivities also differed across generated source populations. At the final review boundary, both eligibility policies filled every content cell in every evaluable form, yet they presented different wording. Across embedding configurations, inclusive primary forms shared a median of only 6 of 40 items, reflecting the total downstream consequence of changing representation across structural evidence and ranking. The apparent stability of global summaries and complete forms therefore concealed instability in the content reaching psychometricians. The computational evaluator is not neutral infrastructure between generation and expertise; it is an inspectable and revisable part of measurement design.

\end{abstract}

\noindent\textbf{Keywords:} 
AI-assisted item generation; computational psychometrics; text embeddings; item reduction; content representation

\hypertarget{introduction}{%
\section{Introduction}\label{introduction}}

Large language models (LLMs) make it inexpensive to generate hundreds or thousands of plausible self-report items. This capacity changes the early stages of scale development because item production can now outpace psychometricians' ability to evaluate what has been produced. A fluent first-person item may appear relevant to a target construct while remaining repetitive, overly evaluative, dependent on a narrow setting, or better aligned with a neighboring construct. When generation is abundant, the central problem is how psychometricians decide which generated items deserve their attention and which computational evidence should inform that decision.

Automated screening offers a practical intermediate layer between generation and expert review. Text embeddings can encode item language for comparisons with alternative construct descriptions and other items. Network methods can organize large pools and identify possible redundancy, while explicit selection rules can compress those pools into candidate forms intended for expert review. Together, these methods make large candidate pools tractable while creating a consequential evaluator: the representation, structural method, retention rule, and form-construction procedure each determine which content psychometricians ever see and which evidence is available when they judge it.

Computational psychometrics joins data-driven computational methods with theory-driven psychometric reasoning \cite{vonDavier2017ComputationalPsychometrics}, making the relation between method and measurement central. Research using multiverse and specification-curve analyses similarly shows that defensible analytic decisions can yield meaningfully different results from common source material \shortcite{simonsohnEtAl2020SpecificationCurve,steegenEtAl2016Multiverse}. The studies we describe here apply the underlying transparency principle to AI-assisted item development. We hold source populations fixed, vary a bounded set of theoretically relevant evaluator choices, and examine whether those choices change the evidence or content available for review by psychometricians. Within this bounded design, our purpose is to expose consequential decisions in an AI-assisted item-development workflow, because those decisions determine which evidence and items reach psychometricians.

Generated items do not enter this evaluator as neutral raw material. Through its learned parameters and attention over the supplied context, a language model conditions each next-token distribution on the prompt, construct information, and adaptive history available to it. This implicit population-forming process makes some candidate items more likely than others before any explicit screening occurs. Parsing, the generation gate, and quota filling then apply a second, explicit selection. The generator and generation package therefore help determine the source population on which semantic representation, structural evaluation, and candidate-form construction operate.

AI-GENIE provides the immediate methodological foundation for this work \shortcite{russellLasalandraEtAl2026AIGENIE}. It connects adaptive item generation with embeddings, Exploratory Graph Analysis (EGA), Unique Variable Analysis (UVA), bootstrap stability evidence, and item reduction. This integration demonstrates how generation and computational reduction can be joined into a coherent pathway for producing candidate forms for psychometric review. It also creates an opportunity to ask a different question: when representation, structural, and candidate-form choices vary, what evidence and content does the evaluator make observable?

Each component contributes to the answer. The generator and generation package first shape which candidate items are produced, which enter the eligible item pool, and which quota-selected items form the source population. An embedding configuration then supplies the item-to-item relations on which structural evaluation depends. Through retention and removal, a structural method changes the population available for review. A candidate-form eligibility policy decides which retained items enter the form-eligible item pool, and joint assignment determines which of those items occupy primary or alternate positions. Reasonable choices at each stage may therefore preserve some properties while changing others.

The distinction between stability of form and stability of content is especially important. A candidate form can fill every content cell and every primary or alternate position while changing many of the items that fill those positions. A final community structure can show stronger correspondence with the intended attribute assignments while losing every item assigned to one attribute. Aggregate summaries of coherence or completeness can conceal these changes.

Throughout this work we use the term candidate form for a computationally assembled set of items presented for expert review. A candidate form is a practical review object that concentrates attention while preserving the uncertainty and evidence that produced it, and it precedes the construction of a final instrument. Thus our studies examine evaluator behavior before response-data validation. Item functioning, score reliability, respondent dimensional structure, and validity in use remain questions for future expert and empirical evaluation.

We investigate the evaluator through two connected studies. Study 1 uses the strongest experimental control in the paper: two generators and two generation packages create source populations that enter five embedding configurations and two structural methods. It asks whether representation changes construct alignment, semantic organization, structural evidence, item retention, and attribute coverage. Study 2 then compares two routes from structural evidence to candidate forms: an inclusive policy accepts an item retained by either method, whereas an agreement policy requires both methods to retain it, and asks whether the resulting candidate forms preserve the same content coverage and wording.

Together, the studies address one overarching question:

\begin{quote}
How do generation, representation, structural screening, and candidate-form construction shape the evidence and content that become available for expert judgment?
\end{quote}

Our contribution is to show why the computational evaluator must itself become an object of psychometric scrutiny. Across fixed source populations, seemingly technical choices in representation, structural reduction, and form eligibility changed the construct evidence attached to identical items, the content that remained available, and the exact wording assembled for expert review, even when broad semantic geometry or form completeness appeared stable. Those consequences could not be recovered from global geometry, recovered community count, or completed quotas alone, because none identifies which items or intended content changed. By preserving source-population identity, task-specific item identities, paired configurations, and stage-specific decisions, we locate where each difference enters and what it alters. Psychometric scrutiny of a candidate form therefore cannot stop at coherence and completeness; it must also examine the evaluator choices and item-level trajectories that produced it.

\hypertarget{conceptual-and-empirical-background}{%
\section{Conceptual and Empirical Background}\label{conceptual-and-empirical-background}}

\hypertarget{ai-assisted-item-generation-and-the-evaluation-bottleneck}{%
\subsection{AI-Assisted Item Generation and the Evaluation Bottleneck}\label{ai-assisted-item-generation-and-the-evaluation-bottleneck}}

Automatic item generation has long treated item production as a design problem rather than unconstrained writing. Classical item models specify the structure that generated tasks or items should inherit, while evidence-centered design begins with the claims and evidence that an assessment is meant to support \shortcite{gierlLai2012ItemModels,mislevyEtAl2003ECD}. Scale-development guidance likewise treats content definition, representative sampling, item wording, and subsequent empirical evaluation as connected parts of instrument development \cite{clarkWatson2019ConstructingValidity,hinkin1998MeasureDevelopment,simms2008ScaleConstruction}. These traditions keep an explicit account of what an item should represent at the center of efficient production.

Contemporary language models extend item generation into domains that are less readily governed by templates. Recent work has used language models to produce construct-specific items, predict item relations, balance content, and support expert or empirical refinement \shortcite{hernandezNie2023AIIP,hoffmannEtAl2024AIScaleDevelopment,hommelEtAl2022AutomaticItemGeneration}. These studies show that language models can contribute to scale development, but inexpensive generation also redistributes the evaluation bottleneck across later decisions. When generation becomes inexpensive, selectivity is distributed across the model's conditional generation, the generation gate, and the later evaluator stages that screen and present the resulting population.

AI-GENIE is useful for the present question because it joins generation and structural reduction while leaving consequential evaluator choices available for inspection. We therefore separate generation eligibility, semantic representation, structural evaluation, and form eligibility so their consequences can be followed from a fixed source population to the candidate form.

\hypertarget{semantic-evidence-and-representation-dependence}{%
\subsection{Semantic Evidence and Representation Dependence}\label{semantic-evidence-and-representation-dependence}}

Item language contains psychometrically relevant information. Semantic models and embeddings have recovered aspects of questionnaire structure \shortcite{milanoEtAl2025SemanticAnalysis,wulffMata2025SemanticEmbeddings}, predicted empirical relations among psychological items and scales \cite{hommelArslan2025LanguageModels,wulffMata2025SemanticEmbeddings}, and mapped overlap among measures and constructs \cite{hommelArslan2025LanguageModels,wulffMata2025SemanticEmbeddings}. Comparisons with human content judgments provide a further form of evidence about item-to-construct alignment \shortcite{milanoEtAl2026ContentValidity}. This literature supports the use of text representations as evidence for screening, although the meaning of that evidence depends on the comparison being made.

Similarity to a target can arise for different reasons. An item may name the construct directly, express a diagnostic behavior, repeat language from a supplied definition, or share a general evaluative tone. Raw cosine values retain the scale of their separately trained embedding systems. The useful question is therefore comparative: whether the intended target is closer than a plausible alternative in the comparison set, whether that ordering changes across representation spaces, and whether those differences propagate into structural evaluation and candidate-form construction.

Representation extends beyond anchor scores. An embedding configuration determines every item-to-item relation supplied to a structural method. Two configurations can preserve a similar overall ordering of item-pair similarities while differing at the local edges or neighborhoods that govern a community, a redundancy decision, or a narrow rank cutoff. Geometry agreement can therefore coexist with changes in item retention. A semantic representation should be evaluated through the consequences it produces at the evaluator stage where it is used.

\hypertarget{reduction-construct-representation-and-content-preservation}{%
\subsection{Reduction, Construct Representation, and Content Preservation}\label{reduction-construct-representation-and-content-preservation}}

Structural evaluation locates two sources of reduction and compares community structure before and after that reduction. Exploratory Graph Analysis \citewithlabel{EGA}{golinoEpskamp2017EGA} provides the community estimates. Unique Variable Analysis \shortcitewithlabel{UVA}{christensenEtAl2023UniqueVariable} first identifies local dependence or redundant variables, and bootstrap EGA evaluates item stability among the remaining items \cite{christensenGolino2021BootstrapEGA}. After unstable items are removed, EGA estimates the final community structure of the retained items. We compare this structure with the corresponding community structure in the full source pool. Together, these procedures organize and reduce a source population; however, the final community structure does not establish whether the intended content survived the reduction. It may correspond more closely to the intended attribute assignments while the retained set no longer includes any item for one attribute. Highly homogeneous sets can similarly preserve coherence by narrowing the content represented \cite{boyle1991ItemHomogeneity}. Community correspondence, exact item identity, and content coverage therefore require separate evidence.

Validity theory supplies the reason for preserving these distinctions. Content evidence concerns the adequacy and relevance of the represented domain, while discriminant content validity asks whether an item represents its intended construct rather than another plausible interpretation \shortcite{aeraEtAl2014Standards,dixonJohnston2019ContentValidity,haynesEtAl1995ContentValidity}. Setting, referent, role, and opportunity can alter what an item represents even when its apparent topic remains target-relevant \shortcite{widhiarsoEtAl2025ConstructIrrelevant}. A computational evaluator can compare only the distinctions represented in its rules, anchors, and structural evidence.

The policy that converts structural evidence into form eligibility adds another layer of judgment. Requiring corroboration from multiple methods may appear conservative, but it can also remove construct-relevant content that one method retained. Accepting support from either method preserves a larger form-eligible item pool, but a larger pool changes the competition for finite form positions. Neither policy is self-justifying, and their consequences must be examined in the candidate forms they produce.

\hypertarget{methods}{%
\section{Methods}\label{methods}}

\hypertarget{study-program-and-common-design}{%
\subsection{Study Program and Common Design}\label{study-program-and-common-design}}

\begin{figure}
\hypertarget{fig:evaluator}{%
\centering
\includegraphics[width=\textwidth,height=0.68\textheight,keepaspectratio]{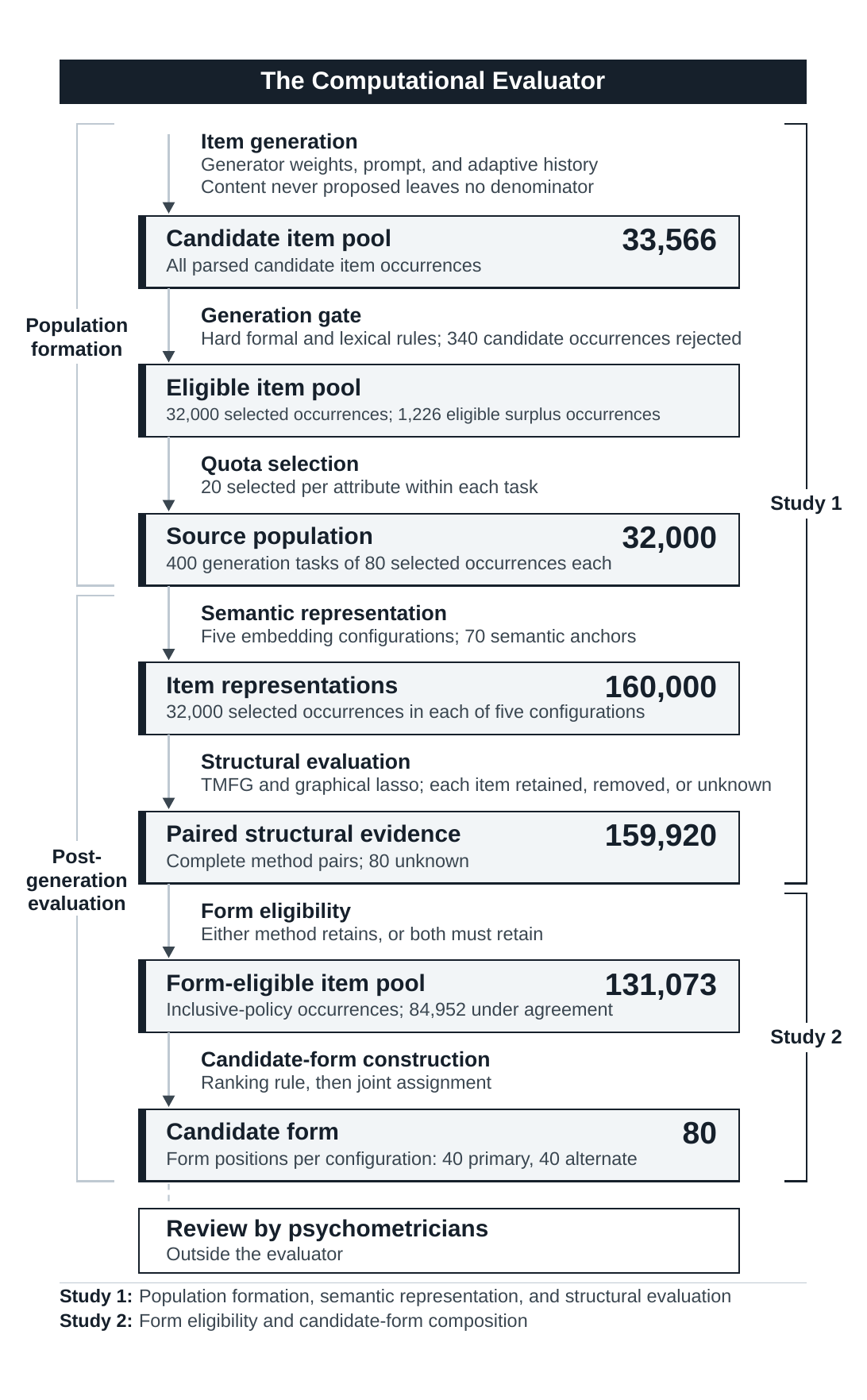}
\caption{The two-study computational evaluator. Generation and quota selection form source populations; semantic representation and paired structural evaluation produce evidence for inclusive or agreement form eligibility; joint assignment then constructs candidate forms. Counts change from item occurrences before embedding to item-by-embedding occurrences after representation and form positions at the endpoint. Study 1 spans source-population formation through semantic representation and structural evaluation, Study 2 follows that evidence into candidate forms, and psychometric review lies outside the evaluator.}\label{fig:evaluator}
}
\end{figure}

The two studies follow the path from a candidate item pool to the content assembled for expert review. Study 1 spans source-population formation through semantic representation and structural evaluation. Study 2 examines the policy that converts paired structural evidence into a candidate form. Figure~\ref{fig:evaluator} shows these shared stages, and Table~\ref{tab:1} summarizes the role of each study.

We use the Big Five as a familiar and conceptually rich test bed. Each of the five traits is represented by four attributes, yielding 20 trait/attribute content cells, such as Openness/curious, that organize generation, semantic anchors, coverage summaries, and candidate-form positions. These attributes serve as working targets for the evaluator; Appendix A provides their definitions, behavioral indicators, and principal boundaries.

The intended response process is an adult rating a self-descriptive personality item on a Likert agreement scale. Item wording was expected to describe a disposition in first-person language without response options or list structure. The construct framework supplied trait and attribute definitions, behavioral indicators, and principal boundaries and confounds that delimited the working targets. Attribute definitions and behavioral indicators served as declared semantic comparisons and, under the construct-indirect package, as generation guidance. This made the evaluator's representation of each target inspectable. The shared endpoint is a candidate form for judgment by psychometricians, not a finalized instrument or a form administered to respondents.

\Needspace{10\baselineskip}
\begingroup
\setlength{\tabcolsep}{2.00pt}
\renewcommand{\arraystretch}{1.18}
\begin{longtable}[]{@{}
  >{\raggedright\arraybackslash}p{(\columnwidth - 6\tabcolsep) * \real{0.1800}}
  >{\raggedright\arraybackslash}p{(\columnwidth - 6\tabcolsep) * \real{0.3500}}
  >{\raggedright\arraybackslash}p{(\columnwidth - 6\tabcolsep) * \real{0.2500}}
  >{\raggedright\arraybackslash}p{(\columnwidth - 6\tabcolsep) * \real{0.2200}}@{}}
\caption{Questions and contributions of the two studies}\label{tab:1} \\
\toprule\noalign{}
\begin{minipage}[b]{\linewidth}\raggedright
Study
\end{minipage} & \begin{minipage}[b]{\linewidth}\raggedright
Primary question
\end{minipage} & \begin{minipage}[b]{\linewidth}\raggedright
Design
\end{minipage} & \begin{minipage}[b]{\linewidth}\raggedright
Contribution
\end{minipage} \\
\midrule\noalign{}
\endfirsthead
\toprule\noalign{}
\begin{minipage}[b]{\linewidth}\raggedright
Study
\end{minipage} & \begin{minipage}[b]{\linewidth}\raggedright
Primary question
\end{minipage} & \begin{minipage}[b]{\linewidth}\raggedright
Design
\end{minipage} & \begin{minipage}[b]{\linewidth}\raggedright
Contribution
\end{minipage} \\
\midrule\noalign{}
\endhead
\bottomrule\noalign{}
\endlastfoot
Study 1: Semantic representation and structural evaluation & How much do construct evidence, item retention, content coverage, and recovered community structure change when fixed items are encoded and evaluated under different configurations, and does that sensitivity differ across source populations? & Identical items compared within each of two generators and two generation packages across five embedding configurations and two structural methods & Locates dependence introduced by representation and structural reduction and tests whether it depends on upstream population formation \\
Study 2: Form eligibility and candidate forms & Does requiring agreement between structural methods preserve the same material for review as accepting support from either method? & Matched inclusive and agreement forms built with a common ranking and assignment procedure & Separates stability of content-cell coverage from stability of exact item wording \\
\end{longtable}
\endgroup

\hypertarget{generators-generation-packages-and-repetitions}{%
\subsection{Generators, Generation Packages, and Repetitions}\label{generators-generation-packages-and-repetitions}}

Item generation crossed two similarly scaled, open-weight generators, Qwen3.5-27B \cite{qwenTeam2026Qwen35} and Gemma 3 27B-IT \cite{gemmaTeam2025Gemma3}; two generation packages; five traits; and 20 independently seeded repetitions. The generators supplied separately realized source populations through which we could examine whether conclusions about the evaluator repeated across model families. Their comparison does not isolate architecture, training, or another model property and is not a ranking of generator quality. We define a generation task as one adaptive sequence specified by generator, generation package, trait, and repetition. Each task produced a source population of 80 selected items, balanced across the four attributes for a given trait. The complete design yielded 400 generation tasks and 32,000 selected items.

The two generation packages differed in the construct information and lexical constraints they supplied. The AI-GENIE source package used compact trait and attribute names and permitted those terms in candidate items. The construct-indirect package supplied attribute definitions, behavioral indicators, and item-writing guidance, while its generation gate prohibited the declared trait and attribute terms in respondent-facing items. Each package therefore combines prompt content, adaptive history, and gate rules, and the comparison concerns the complete generation packages rather than a prompt-only or gate-only effect. Appendix A reproduces the complete package instructions and construct framework.

Within each request, the generator conditioned its response on the package's construct information, instructions, and current adaptive history. Its learned representation and attention over that context shaped which candidate items were proposed; the package-specific generation gate then determined which could enter the eligible item pool. Quota selection from that pool produced the source population. The source population was therefore the product of both implicit model conditioning and explicit eligibility decisions.

Requests within a task were adaptive, with recent eligible items returned to later requests to discourage repetition; items filled attribute quotas in request order. Items produced within a task therefore shared an evolving adaptive history, making the task the repeated unit for resampling and uncertainty summaries. A bounded two-trait calibration compared nested source populations of 40, 60, 80, and 100 items and supported the choice of 80 items as a pragmatic source-population depth; it did not establish a generally optimal pool size (Appendix E).

\hypertarget{semantic-representation-and-structural-evaluation}{%
\subsection{Semantic Representation and Structural Evaluation}\label{semantic-representation-and-structural-evaluation}}

Each source population was encoded separately by five embedding configurations: Qwen3-Embedding-0.6B, Qwen3-Embedding-4B, Qwen3-Embedding-8B, EmbeddingGemma-300M, and BGE-M3. Appendix B reports their vector dimensions, input modes, and design roles. The configurations differ jointly in model family, training, architecture, parameter count, dimension, and input treatment. The Qwen configurations provide a bounded within-family comparison, while BGE-M3 and Qwen3-Embedding-0.6B provide an informative same-dimension comparison.

The Big Five construct framework produced five families of semantic anchors: trait labels, trait definitions, attribute labels, attribute definitions, and behavioral indicators. Within each family, the intended target was compared with the strongest alternative in the comparison set. Target advantage preserved zero as a tie between the target and its strongest alternative. To compare its magnitude across embedding configurations, we divided target advantage by its standard deviation within each embedding configuration and anchor family without shifting the zero point. Target rank and the target-first rate supplied companion evidence. The purpose of this contrastive design was to make the comparison explicit rather than treating similarity to a target in isolation as construct evidence.

For each generation task and embedding configuration, the common GENIE-compatible structural sequence first used UVA to examine local dependence. After UVA, TMFG \shortcite{massaraEtAl2017TMFG} and graphical lasso \shortcite{friedmanEtAl2008GraphicalLasso} independently evaluated the resulting item pool and produced method-specific bootstrap-stability evidence. The configured EGA procedure treated embedding-coordinate rows as cases, making vector dimension the effective sample size. This value informed graphical-lasso regularization and the parametric stability bootstrap under both methods, whereas TMFG initial-network construction used the correlation matrix without using the sample-size argument. Dimension therefore remained one component of the complete embedding configuration rather than an isolated cause. We compared the community structure of the full 80-item source pool with the community structure of the items retained after UVA and stability reduction. Each item retained a terminal state indicating whether it survived, was removed through UVA, was removed below the stability threshold, or remained unknown. Appendix B reports the evaluator settings, recurring outcome definitions, and form-construction procedure in greater detail.

\hypertarget{item-eligibility-and-evidence-preservation}{%
\subsection{Item Eligibility and Evidence Preservation}\label{item-eligibility-and-evidence-preservation}}

Candidate items passed through the generation gate before entering the source population. The gate required one interpretable self-report item within the declared length range, excluded response-option and list artifacts, removed exact normalized duplicates, and applied the direct-term prohibition under the construct-indirect package. Review warnings identified possible problems such as double-barreled content, negation, nonstandard self-report form, or strongly evaluative wording. Warnings remained attached to eligible items for later review rather than becoming an unvalidated numerical quality penalty.

The evidence record preserved attempted outputs, generation-gate decisions, eligible surplus, primary and alternate candidates, structural trajectories, and unknown evidence. Items retained task-specific identities, so the same normalized wording generated in different adaptive histories remained distinct in the analysis. Normalized wording was used separately to prevent the same text from occupying more than one candidate-form position. This preservation keeps the source population and changing denominators available for interpretation and permits evaluator configurations to be compared without regenerating the items.

\hypertarget{missing-and-non-evaluable-evidence}{%
\subsection{Missing and Non-evaluable Evidence}\label{missing-and-non-evaluable-evidence}}

Missing and non-evaluable evidence remained explicit in denominator accounting rather than being reclassified as negative evidence. One BGE-M3/TMFG structural evaluation ended in a repeatable analytic noncompletion and remained unknown wherever its result was required. Because retuning or replacing the bootstrap stability procedure after observing this result would have changed the evaluator under study, we retained the configured procedure and treated the result as neither retention nor removal evidence. The affected outcomes remained distinct from removal, method disagreement, and an empty form.

\hypertarget{study-1-representation-and-structural-evidence}{%
\section{Study 1: Representation and Structural Evidence}\label{study-1-representation-and-structural-evidence}}

\hypertarget{purpose-and-research-question}{%
\subsection{Purpose and Research Question}\label{purpose-and-research-question}}

Study 1 asks what changes when identical generated items are represented in different embedding spaces and then evaluated through the same structural methods. Fixed-item comparisons provide an important control: within each source population, any difference begins after generation and can be located in the later evaluator stages. That control does not make generation irrelevant, and comparing the same representation contrasts across generators and generation packages tests whether evaluator sensitivity depends on the population entering it.

\begin{quote}
RQ1. For fixed source populations from each generator and generation package, how much do construct evidence, item retention, content coverage, redundancy behavior, semantic geometry, and recovered community structure vary across embedding configurations and structural methods, and does that variation differ by generator or generation package?
\end{quote}

The complete five-embedding panel is the primary comparison, and the three Qwen configurations provide a secondary within-family comparison, while the two 1,024-dimensional configurations allow a limited test of whether equal vector dimension is sufficient to produce similar evidence. Generator and generation-package contrasts ask whether representation sensitivity recurs across independently realized source populations; they do not rank the generators or separate the individual prompt and generation-gate components of a package. Because each embedding configuration bundles several model properties, the study describes representation sensitivity across complete configurations.

\hypertarget{construct-evidence-and-outcomes}{%
\subsection{Construct Evidence and Outcomes}\label{construct-evidence-and-outcomes}}

Semantic evidence was evaluated through declared contrasts. For every anchor family, the intended trait or attribute was compared with the strongest alternative represented in the same family. This preserved a meaningful zero point and made the identity of the competing construct visible. Label anchors showed how the embeddings treated direct lexical cues, whereas definitions and behavioral indicators provided richer descriptions of the intended content.

Study 1 then explored representation across five additional properties: retained-set overlap asked whether the same source items survived under two configurations; content coverage asked whether each of the four intended attributes remained represented after reduction; stage-specific removal distinguished UVA from later stability-based decisions; geometry agreement summarized the ordering of all item pairs within a task; and community correspondence described how closely the recovered communities aligned with the intended attribute assignments. We kept these outcomes separate because a stable count or a simpler community structure could coexist with different items or missing content.

Because every embedding configuration received the same source population, Study 1 paired comparisons within generation task. Shared bootstrap draws resampled generation tasks within each combination of generator, generation package, and trait. We then compared task-level sensitivities across generators and across generation packages. Different generator and generation-package cells contained independently generated items, so these cross-population contrasts compare sensitivity estimates rather than item-level pairs. The resulting intervals describe variation across repeated generation tasks. Interpretation emphasizes effect magnitude, overlap, coverage, and the location of disagreement at the task level; participant-sampling uncertainty lies outside this design. The complete fixed-design analysis was computed before a bounded set of results was selected for manuscript display. Displays were chosen after inspection to answer the study's questions, and the intervals remain descriptive summaries of task-level variation.

\hypertarget{results}{%
\subsection{Results}\label{results}}

The five embedding configurations attached different construct evidence to the same items, and the difference depended on how the construct was represented (Appendix C, Tables~\ref{tab:C1}--\ref{tab:C3}). Qwen 8B produced the largest average scaled target advantage across all five anchor families. For behavioral indicators, however, Qwen 8B and Qwen 4B were nearly tied: their scaled target advantages were 1.417 and 1.408, and their target-first rates were 92.16\% and 92.52\%, respectively. That near tie did not mean that the same items received the same evidence: target-first status varied across configurations for 35.2\% of AI-GENIE source items and 25.5\% of construct-indirect items. Literal trait labels were the weakest anchors, while definitions and behavioral indicators more clearly separated intended content from its alternatives. Representation choice therefore affected not only the average strength of construct evidence, but also which particular statements appeared to fit their intended targets.

The bounded within-Qwen comparison made this unevenness more explicit. Mean initial adjusted mutual information (AMI) increased across Qwen 0.6B, 4B, and 8B under TMFG (.653, .691, and .715) and graphical lasso (.758, .782, and .791); both adjacent contrasts and the endpoint contrast were positive (Appendix C, Table~\ref{tab:C4}). Yet only 59 of the 100 construct-evidence cells defined by generator, generation package, trait, and anchor family were monotonic across the three levels. Final AMI instead peaked at Qwen 4B within the Qwen comparison: .906 compared with .899 for Qwen 8B under TMFG, and .891 compared with .886 under graphical lasso. The configured sequence therefore produced an ordered aggregate pattern at the initial structural stage, not a uniform larger-is-better result across the evaluator.

The two generation packages also produced different semantic profiles. The AI-GENIE source population aligned more strongly with trait labels, trait definitions, and attribute labels, whereas the construct-indirect population aligned more strongly with behavioral indicators. Under Qwen 8B, the construct-indirect population's behavioral-indicator target advantage was .222 higher than that of the AI-GENIE source population (95\% task-bootstrap interval {[}.202, .242{]}). This pattern is consistent with the language made available by each package: one permitted direct construct terms, while the other supplied behavioral descriptions and excluded those terms from respondent-facing items. Because the construct-indirect attribute definitions and behavioral indicators were also used as anchors, these contrasts show that the package manipulation changed the source population rather than independently confirming construct quality.

An item-to-item analysis supported a related but distinct conclusion. Within each task and embedding, construct-indirect populations showed a larger difference between within-attribute and between-attribute similarity across all ten generator-by-embedding comparisons (Appendix C, Table~\ref{tab:C2}). Because the calculation compared relations among generated items rather than items with anchors, it did not reuse the anchor panel. More separated source populations sometimes produced more fragmented networks, showing that average semantic organization did not determine the local structural relations recovered later.

Embedding sensitivity also differed by generator. Across all ten embedding pairs and both structural methods, Qwen-generated source populations produced greater initial community correspondence and greater raw retained-set overlap than Gemma-generated populations; the overlap contrast combines differences in retention volume and exact retained identity. For the Qwen 4B-Qwen 8B comparison under TMFG, the differences were .080 in retained-set Jaccard (95\% task-bootstrap interval {[}.054, .106{]}) and .065 in initial community AMI ({[}.048, .083{]}), meaning that the two representations preserved more of the same items and more similar initial partitions in the Qwen-generated populations. Overall geometry differences were smaller and changed direction across pairs, so the moderation appeared in downstream local decisions rather than in a uniform difference in broad semantic similarity. Qwen-generated items also showed greater behavioral-indicator target separation under all five embeddings, although that direction did not extend to every anchor family. For computational psychometricians, the consequence is that sensitivity observed in one generator's source population cannot be assumed to transfer unchanged to another's.

The generation package changed where representation sensitivity appeared. Under Qwen 4B and Qwen 8B with TMFG, construct-indirect populations began with more recovered communities and lower correspondence with the intended attributes than AI-GENIE-source populations, despite greater average semantic separation among their assigned attribute groups. A localized post hoc contrast made this dependence especially visible for Neuroticism/anxious content: Qwen 8B retained 46.3 percentage points more than BGE-M3 under AI-GENIE source but only 0.8 points more under construct indirect; the corresponding Gemma differences were 31.8 and 2.3 points. This example does not establish that Neuroticism/anxious content is uniquely sensitive or that either package is generally preferable; however, it demonstrates that a generation package can amplify or attenuate a later representation contrast.

Individual items make this contrastive evidence easier to interpret. An Agreeableness/compassionate item about being affected by another person's distress favored its intended attribute under every embedding, while a Conscientiousness/disciplined item describing deliberation before action favored Conscientiousness/prudent throughout the panel. An Agreeableness/humble item about acknowledging another person's superior skill changed its leading attribute: Qwen 0.6B, Qwen 4B, and Qwen 8B placed humble first, whereas EmbeddingGemma and BGE-M3 placed compassionate first. These are purposive illustrations rather than prevalence estimates, but they show how fixed wording can acquire a different construct ordering when its representation changes.

Representation changed the exact statements that survived: under TMFG, average retained-set overlap across embedding pairs ranged from .410 to .565. Even Qwen 4B and Qwen 8B, the closest pair in this comparison, shared only 56.5\% of the union of their retained items. Graphical lasso preserved larger populations and produced higher overlap, ranging from .666 to .755, but still retained distinct sets. The direction or quality of these substitutions requires review by psychometricians. What the evaluator makes visible here is their scale: representation choice changed many of the statements available for judgment, and failing to inspect that change would mistake a configured result for a neutral one (Appendix C, Table~\ref{tab:C5}).

Balanced generation was followed by uneven survival. Every task began with 20 items for each of four attributes. After TMFG reduction, average retention ranged from just under one half to nearly two thirds across embeddings. BGE-M3 produced 119 empty content cells among 1,596 evaluable cells, compared with 27 of 1,600 under Qwen 8B. Graphical lasso retained more content overall and rarely produced an empty cell, although its coverage still varied across representations. At the trait level, the Openness trait was most configuration-sensitive, whereas Conscientiousness varied less across the embedding panel but produced the most empty attribute cells. Thus a comparatively stable average retention rate did not protect every intended content area; coverage had to be examined separately from total count.

Stage-specific removal evidence showed why a single endpoint count was insufficient. In several large contrasts, the configuration retaining more items had removed slightly more through UVA but substantially fewer during stability reduction. For Openness/creative content under TMFG, for example, Qwen 8B's retention advantage over BGE-M3 arose during the stability stage rather than from uniformly less selective treatment. Locating each removal at the stage where it occurred distinguishes local-dependence evidence from later instability, identifies which stage warrants scrutiny, and shows that changing an endpoint requires changing a particular decision rule rather than treating the evaluator as a single black box.

The evidence converged across layers: representation altered construct comparisons, retained identity, coverage, and recovered structure, while the source population changed the magnitude of those effects, as shown in Table~\ref{tab:2}.

\begin{table}[!tb]
\caption{Linked Study 1 evidence from representation to retained content. Brackets contain equal-tailed 95\% task-bootstrap intervals for selected task-level comparisons; Table~\ref{tab:C6} in Appendix C reports their task counts and supporting evidence boundaries.}\label{tab:2}
\centering
\begingroup
\scriptsize
\setlength{\tabcolsep}{4.50pt}
\renewcommand{\arraystretch}{1.10}
\begin{tabular}{@{}
  >{\raggedright\arraybackslash}p{(\columnwidth - 6\tabcolsep) * \real{0.2100}}
  >{\raggedright\arraybackslash}p{(\columnwidth - 6\tabcolsep) * \real{0.1900}}
  >{\raggedright\arraybackslash}p{(\columnwidth - 6\tabcolsep) * \real{0.3100}}
  >{\raggedright\arraybackslash}p{(\columnwidth - 6\tabcolsep) * \real{0.2900}}@{}}
\toprule\noalign{}
\begin{minipage}[b]{\linewidth}\raggedright
Evidentiary layer
\end{minipage} & \begin{minipage}[b]{\linewidth}\raggedright
Unit
\end{minipage} & \begin{minipage}[b]{\linewidth}\raggedright
Selected result
\end{minipage} & \begin{minipage}[b]{\linewidth}\raggedright
Interpretive consequence
\end{minipage} \\
\midrule\noalign{}
Construct evidence & 32,000 fixed items across 400 generation tasks & Qwen 8B led all five anchor families in scaled target advantage; for behavioral indicators, Qwen 8B and Qwen 4B had scaled target advantages of 1.417 and 1.408, target-first rates of 92.16\% and 92.52\%, and mean target ranks of 1.095 and 1.096 & Representation changes the construct evidence attached to identical items \\
Retained identity & Paired generation tasks under the same structural method & Retained-set Jaccard ranged from .410 to .565 under TMFG and from .666 to .755 under graphical lasso & Similar endpoint counts can conceal different items \\
Intended coverage & Four attributes within each structural evaluation & TMFG produced substantial differences in empty content cells across embeddings, including 119 under BGE-M3 and 27 under Qwen 8B & Balanced generation can end with an intended attribute unavailable for review \\
Overall geometry and local outcomes & 400 pair-complete generation tasks & Qwen 4B and Qwen 8B had geometry agreement of .874 {[}.872, .875{]}, with TMFG retained-set Jaccard of .565 {[}.552, .578{]} and initial community AMI of .720 {[}.712, .729{]} & Overall semantic agreement can coexist with different networks and retained items \\
Recovered community structure & 3,999 evaluable structural evaluations & Initial AMI was highest for Qwen 8B under both methods (.715 for TMFG and .791 for graphical lasso); exact four-community solutions rose from 32.93\% initially to 73.02\% finally, while AMI decreased in 276 evaluations & Initial correspondence compares complete source populations; final correspondence and change describe retained populations and do not establish a validity gain \\
Source-population dependence & Independently generated 200-task source populations & For Qwen 4B-Qwen 8B under TMFG, Qwen-generated source populations exceeded Gemma-generated source populations by .080 {[}.054, .106{]} in raw retained-set Jaccard and .065 {[}.048, .083{]} in initial community AMI & Representation sensitivity is partly a property of the source population entering the evaluator \\
\bottomrule\noalign{}
\end{tabular}
\endgroup
\end{table}

The embeddings agreed more about the overall ordering of item-pair similarities than about which items the evaluator ultimately retained. Across all embedding pairs, task-level geometry agreement ranged from .733 to .874. The highest geometry agreement occurred between Qwen 4B and Qwen 8B, yet under TMFG their retained-set overlap was only .565 and their initial community AMI was .720. Two representations could therefore preserve much of the same broad semantic ordering while differing at the local relations that changed network boundaries and retention decisions.

We use initial AMI as the leading community-correspondence result because it compares all 80 items in each source population. Qwen 8B led under both structural methods, with initial AMI of .715 for TMFG and .791 for graphical lasso, and initial ARI and NMI provided concordant aggregate evidence. Final AMI and final-minus-initial change instead describe separately retained populations: AMI increased on average but decreased in 276 of 3,999 evaluable structural evaluations. Exact four-community solutions rose from 32.93\% initially to 73.02\% finally, yet four communities could still mix intended attributes or be estimated after one attribute had disappeared. Table~\ref{tab:2} retains this central cross-stage result; Table~\ref{tab:C7} in Appendix C and its accompanying text report the full configuration profiles, method-specific decreases, and NMI/AMI/ARI agreement.

We explored a same-dimension comparison between BGE-M3 and Qwen 0.6B that ruled out the simple explanation that embedding dimensionality alone accounted for the observed differences. The two supplied the same dimension-derived effective sample size to the structural method, yet produced the lowest TMFG retained-set overlap in the panel and differed in initial community correspondence. Equal vector dimension was therefore insufficient to produce equivalent evidence, while architecture, training, geometry, input treatment, and other properties remained bundled within the configurations.

\hypertarget{contribution}{%
\subsection{Contribution}\label{contribution}}

By following fixed items across construct comparisons, local networks, and final retention, Study 1 shows that representation is a condition of the evidence rather than an invisible implementation choice. The same items acquired different target-versus-competitor relations, entered different local networks, survived at different rates, and sometimes left different attributes available. The magnitude and location of those differences also changed with the generator and generation package that produced the source population. Source populations were therefore not interchangeable replications; they moderated what later representation and reduction choices did.

This interaction makes the evaluator divisible into researchable decisions. Computational psychometricians can hold a source population fixed while varying an anchor panel, embedding configuration, or structural method, then locate which relations, removals, coverage losses, and retained statements change. They can also hold the later evaluator fixed while comparing source populations produced by different generators or generation packages. These comparisons do not identify a preferred configuration on their own. They locate where a change first enters the evidence path and which downstream material it alters, allowing follow-up studies to isolate one evaluator decision, test a revision against a fixed source population, and determine whether the intended content is better preserved.

Study 1 also separates structural simplification from content preservation. Reduction often produced a final community structure that corresponded more closely to the intended attribute assignments, but that simpler organization sometimes described fewer items and omitted an intended attribute. Higher final community correspondence and the expected number of communities therefore leave content preservation as a separate empirical question. Study 2 carries this distinction forward by asking how paired structural outcomes become a candidate form.

\hypertarget{study-2-policy-effects-and-candidate-forms}{%
\section{Study 2: Policy Effects and Candidate Forms}\label{study-2-policy-effects-and-candidate-forms}}

\hypertarget{purpose-and-research-question-1}{%
\subsection{Purpose and Research Question}\label{purpose-and-research-question-1}}

Study 2 begins after both structural methods have evaluated the same source population within an embedding configuration. At that point, a researcher must decide how method-specific evidence becomes form eligibility. We compared two reasonable policies for making that decision: an inclusive policy that made an item form-eligible when either method retained it, and an agreement policy that required both methods to retain the same item. The generated items, semantic representations, structural results, anchor evidence, content cells, and form-construction procedure otherwise remain fixed. The two policies therefore differ only when one structural method retains an item and the other does not.

\begin{quote}
RQ2. Within fixed source populations and embedding configurations, what changes when form eligibility requires agreement between the structural methods rather than support from at least one method, and do those changes alter the candidate items presented for expert review?
\end{quote}

This question treats the response to structural disagreement as a policy rather than as a property of the items themselves. Agreement provides corroboration from the method-specific stability evidence after the shared UVA step, leaving item quality for later judgment by psychometricians. The inclusive policy preserves a larger form-eligible item pool and changes which candidates compete near the form boundary. The comparison is therefore empirical: what does each policy make available, and what does each eventually place before psychometricians?

\hypertarget{candidate-form-construction}{%
\subsection{Candidate-Form Construction}\label{candidate-form-construction}}

Study 2 constructed one candidate form for each configuration defined by generator, generation package, and embedding configuration. The structural policy first determined which items could compete for positions within each of the 20 content cells. Eligible items were then ranked by their combined target advantage for attribute definitions and behavioral indicators. Because the construct-indirect package supplied those same definitions and indicators during generation, the score provided package-aligned prioritization rather than independent evidence of item quality. Under the inclusive policy, support from both structural methods broke only a literal score tie. Warning burden and generation-task provenance supplied later deterministic ordering without becoming general quality scores.

Assignment then proceeded jointly across all content cells rather than one cell at a time. This prevented recurring normalized wording from occupying more than one position and made coverage independent of an arbitrary cell order. Within each configuration, both policies sought two primary and two alternate items per content cell, defining an 80-position review object that preserved alternatives for later judgment. Primary and alternate describe form positions, not ordinal ranks among all candidates in a cell. We compared 20 matched inclusive and agreement configurations; one full-data pair was non-evaluable because of the structural noncompletion. Secondary method-specific forms identified which method supplied the additional content under each policy, while paired resampling tested whether policy differences persisted across alternative mixtures of generation tasks.

\hypertarget{results-1}{%
\subsection{Results}\label{results-1}}

Each complete paired item outcome fell into one of four categories: retained by both methods, retained only by graphical lasso, retained only by TMFG, or retained by neither. Slightly more than half (53.1\%) were retained by both methods. About one quarter (25.0\%) were retained only by graphical lasso, a much smaller share (3.8\%) only by TMFG, and the remainder (18.0\%) by neither. The two policies differed only for the one-method outcomes. These accounted for 28.8\% of complete pairs, and most reflected graphical-lasso-only support.

That disagreement substantially changed form eligibility. Requiring both methods to retain an item removed 46,121 of the 131,073 complete item occurrences eligible under the inclusive policy, contracting the pool by 35.2\%. Across the 19 configurations carried through complete form construction, agreement removed from roughly one fifth to more than one half of the inclusive pool. A policy expressible in one sentence therefore changed a substantial portion of the material permitted to compete for review positions.

Items retained by neither method did not distinguish the policies because both policies excluded them. Of these 28,847 occurrences, 15,546 (53.9\%) had already been removed during the shared UVA stage in both traces; the remaining 13,301 survived UVA but were removed during both methods' stability evaluations. This breakdown locates why items were unavailable, but it does not contribute to the 35.2\% policy contraction.

The smaller agreement-eligible pool did not prevent either policy from completing its forms. Every evaluable form under both policies filled all 40 primary and 40 alternate positions, preserving all 20 content cells at the observed pool sizes. Exact wording nevertheless differed: matched primary forms shared a median of 29 of 40 items, leaving a median 11 items unique to each policy. Including alternates increased the difference, with a median 21 items in each policy's form absent from the other. Table~\ref{tab:3} places this policy-driven difference beside the larger divergence associated with changing the embedding configuration.

\Needspace{14\baselineskip}
\begingroup
\setlength{\tabcolsep}{2.00pt}
\renewcommand{\arraystretch}{1.08}
\begin{longtable}[]{@{}
  >{\raggedright\arraybackslash\hyphenpenalty=10000\exhyphenpenalty=10000}p{(\columnwidth - 8\tabcolsep) * \real{0.2200}}
  >{\raggedright\arraybackslash\hyphenpenalty=10000\exhyphenpenalty=10000}p{(\columnwidth - 8\tabcolsep) * \real{0.2100}}
  >{\raggedright\arraybackslash\hyphenpenalty=10000\exhyphenpenalty=10000}p{(\columnwidth - 8\tabcolsep) * \real{0.1900}}
  >{\raggedright\arraybackslash}p{(\columnwidth - 8\tabcolsep) * \real{0.2200}}
  >{\raggedright\arraybackslash}p{(\columnwidth - 8\tabcolsep) * \real{0.1600}}@{}}
\caption{How structural-method policy and embedding configuration changed candidate-form eligibility and wording. The first two rows follow the policy comparison from eligibility loss to overlap between matched 40-item primary forms. The third compares inclusive forms across embedding configurations, and the fourth summarizes paired generation-task resamples. Shared-primary counts and Jaccard values compare normalized wording; form results show whether all 20 content cells received two primary and two alternate items. Appendix D, Table~\ref{tab:D1} provides the configuration-level evidence.}\label{tab:3} \\
\toprule\noalign{}
\begin{minipage}[b]{\linewidth}\raggedright
Comparison
\end{minipage} & \begin{minipage}[b]{\linewidth}\raggedright
Evaluable evidence
\end{minipage} & \begin{minipage}[b]{\linewidth}\raggedright
Form result
\end{minipage} & \begin{minipage}[b]{\linewidth}\raggedright
Shared primary wording
\end{minipage} & \begin{minipage}[b]{\linewidth}\raggedright
Set agreement
\end{minipage} \\
\midrule\noalign{}
\endfirsthead
\toprule\noalign{}
\begin{minipage}[b]{\linewidth}\raggedright
Comparison
\end{minipage} & \begin{minipage}[b]{\linewidth}\raggedright
Evaluable evidence
\end{minipage} & \begin{minipage}[b]{\linewidth}\raggedright
Form result
\end{minipage} & \begin{minipage}[b]{\linewidth}\raggedright
Shared primary wording
\end{minipage} & \begin{minipage}[b]{\linewidth}\raggedright
Set agreement
\end{minipage} \\
\midrule\noalign{}
\endhead
\bottomrule\noalign{}
\endlastfoot
Inclusive versus agreement form eligibility & 159,920 of 160,000 planned item outcomes had complete paired structural evidence & Agreement removed 46,121 of 131,073 complete inclusive-form-eligible item occurrences (35.2\%) & Not applicable & Not applicable \\
Inclusive versus agreement forms & 19 of 20 matched configurations & Both policies filled every primary and alternate position & Median 29 of 40 & Median Jaccard .569 \\
Inclusive forms across embeddings & 36 of 40 within-population embedding pairs & Compared forms filled the same 20 content cells & Median 6 of 40 & Median Jaccard .081 \\
Inclusive versus agreement task resamples & 19,359 of 20,000 paired draws & Every evaluable draw filled all positions & Varied by configuration & Median Jaccard ranged from .379 to .778 \\
\end{longtable}
\endgroup

Neither policy showed a consistent advantage on the supporting diagnostics. Warning burden changed direction across configurations, selected items under both policies had survived the shared UVA step, and both policies drew primary items from a similarly broad set of source tasks. The demonstrated policy consequence was therefore a change in the items presented for review, not a consistent improvement in warning burden, local-dependence evidence, or source concentration.

The method-specific forms explained why the policies produced this pattern. Because graphical lasso retained more items, inclusive forms closely resembled forms built from graphical-lasso retention alone. Agreement forms, restricted to items also retained by the more selective TMFG procedure, more closely resembled forms built from TMFG retention alone. This relationship explains the realized form differences without establishing that either method produced better items.

We also compared the policy contrast with a broader change to the evaluator. Within one embedding configuration, the inclusive and agreement policies used the same structural results and item ranking; only the eligibility rule changed. Their primary forms shared a median of 29 of 40 items. The cross-embedding comparison held the generated source population fixed but changed the representation used to produce both the structural evidence and the item ranking. Those forms shared a median of only 6 of 40 items. This larger divergence does not rank representation against policy, because the two comparisons differ in scope: the policy contrast changed one final rule, whereas the embedding contrast propagated through several stages of the evaluator. The result instead shows that the representation differences observed in Study 1 remained visible in the exact items presented for review.

Why could forms differ this much when broader rankings remained similar? The divergence concentrated near the candidate-form cutoff rather than reflecting a wholesale reversal of every item ranking. Across all items assigned to the same trait/attribute content cell (defined in Appendix A), such as Conscientiousness/disciplined, embedding pairs showed moderate to high rank agreement and substantial overlap in their form-eligible pools. Yet the four highest-ranked eligible candidates overlapped by only one item on average. A narrow review quota can therefore amplify local ordering differences that appear modest in a global summary.

Paired resampling tested whether the policy result depended on the realized mixture of generation tasks. Every evaluable draw produced a complete form under both policies, and wording differences persisted across the central range of comparisons within each configuration. The configuration affected by the structural noncompletion was evaluable only in draws that excluded the incomplete task, so its resampling results are conditional on complete paired structural evidence.

\hypertarget{contribution-1}{%
\subsection{Contribution}\label{contribution-1}}

Study 2 separates two kinds of stability: whether a form is complete and whether it contains the same wording. At the observed pool sizes, both policies produced complete, balanced forms in all 19 evaluable configurations, but exact wording remained policy-dependent. Requiring corroboration excluded roughly one third of the inclusive form-eligible pool and left a median of 11 of 40 primary items specific to each policy.

This distinction matters because psychometricians judge items, not empty content-cell labels. Filling every cell does not establish that the selected items are suitable for their intended purpose, and reporting only complete quotas would make the evaluator's selection appear neutral. The policies instead presented different wording for expert judgment. The observed cross-embedding divergence completes the progression from Study 1: representation changed structural outcomes, and those differences remained visible when the evaluator selected a small set of items for review.

\hypertarget{general-discussion}{%
\section{General Discussion}\label{general-discussion}}

AI-assisted item development places a sequence of computational choices between construct specification and judgment by psychometricians. The first choices form the source population: a generator conditions candidate production on its learned weights and supplied context, while a generation package defines the prompt, adaptive history, and generation gate. Semantic representation, structural evaluation, and candidate-form construction then determine which items and evidence survive long enough to be judged. The resulting evaluator is useful because no researcher can carefully examine every item in a very large candidate item pool. But the same evaluator is not a neutral part of the process: its methods help determine what reaches expert review. Across two connected studies, we made those decisions inspectable by comparing source populations, holding each population fixed for later contrasts, preserving task-specific item identities and stage histories, and carrying representation and structural differences through to the candidate forms assembled for review.

The findings support a property-specific account of stability: across embedding configurations, much of the overall semantic ordering among fixed items was preserved, reduction usually increased community correspondence with the intended attribute assignments, and both form policies filled every position in every evaluable configuration. These are meaningful forms of stability.

The material available for later review was less stable even when these broader properties held: geometry agreement coexisted with lower overlap among the exact items retained, and community count and intended content coverage were not the same. Four estimated communities did not guarantee that all four intended attributes remained represented: if every item assigned to one attribute disappeared, the surviving items could still divide into four communities because another attribute split across more than one community. Complete candidate-form quotas likewise coexisted with policy-specific wording, and forms built under different embedding configurations shared few exact items. These findings matter because stability in geometry, community count, or quota fill does not ensure stability in the content psychometricians receive.

Study 1 places both source-population formation and semantic representation within the measurement design. A generator is often treated as a convenient way to obtain text, and an embedding as a technical input chosen before analysis begins. In this workflow, the generator's learned weights and the generation package's prompts, adaptive history, and gate rules shaped which items entered the source population. The embedding configuration then determined the construct contrasts attached to those items, the item relations supplied to the network, the community structure recovered from those relations, and the content available after reduction. Even closely related embedding configurations could agree broadly about semantic geometry while differing at the local relations that governed survival, and the size of that sensitivity differed across source populations. What could become evidence therefore depended on both how items entered the evaluator and how the evaluator represented them.

The anchor panel imposed a second representational boundary. A target score answers a comparison defined by the selected target, alternatives, definitions, and indicators. Study 1 showed that the resulting construct evidence changed with both the representation and the anchor family. Psychometricians should therefore treat the comparison set as part of the construct representation rather than as an incidental scoring aid.

This point also clarifies what can be learned from the two generators and two generation packages. The generators supplied separately realized populations and showed whether later representation effects repeated across two model families. Each package changed prompt content, supplied construct language, adaptive history, and generation-gate rules together. Their anchor profiles and later moderation therefore describe complete generation packages rather than isolated effects of instruction wording or label prohibition. Generation changes the source population, whereas a new anchor panel, structural method, or candidate-form policy can be applied to fixed items. These are distinct interventions, but their consequences can interact and should be studied together. As evaluator choices proliferate, treating an available configuration as a neutral default becomes increasingly difficult to defend: the starting population and the later evaluator can each alter what becomes evidence and what reaches psychometricians.

Structural evidence and content evidence were similarly connected without being interchangeable. EGA, UVA, and bootstrap stability characterized the organization, local dependence, and stability of the configured item relations. These procedures often produced stronger correspondence between the final community structure and the intended attribute assignments while content coverage declined. The resulting structural evidence did not by itself establish that every intended attribute remained available for review.

The practical task is to preserve the relationship between structural and construct evidence while keeping their functions distinct. A psychometrician should be able to see an item's declared target, relevant semantic contrasts, structural trajectory under each method, content cell, competing items near the form cutoff, and eventual position. This makes it possible to distinguish an item removed for local dependence from one whose wording raises a construct concern, and to inspect relevant content that the primary form displaced.

Study 2 makes the policy between structural evidence and expert review inspectable. Requiring agreement substantially narrowed the form-eligible item pool, but both policies still produced complete and balanced forms in all 19 evaluable configurations. If the analysis had stopped at form completeness, the policy would have appeared inconsequential. Exact wording showed otherwise. The policies presented different items for judgment, and changing the embedding configuration changed them more. A candidate form can therefore cover all content cells while remaining dependent in its item composition.

The narrow form quota helps explain this dependence. Global rank agreement and overlap between form-eligible item pools remained much higher than overlap among the few items occupying each content cell. Small local changes near a cutoff can decide what is included even when the overall ordering looks similar. This general consequence follows from asking a computational evaluator to compress a large, competitive population into a small review object. Evaluator reports should consequently connect global summaries to exact form composition and the candidates displaced at the form cutoff.

These findings show why psychometricians need a handoff that reveals how the review set was formed. Candidate forms can concentrate attention while preserving their relationship to the source population. Review material can retain a primary and alternate form while also identifying method disagreements, small or negative target advantages, attributes vulnerable to loss, and high-ranking candidates displaced near a quota boundary. The broader principle is that psychometricians should receive the evidence needed to understand how a finite review set came to exist.

Traceability matters because discarded content and stage histories cannot be reconstructed from a final item list. Once they disappear, later reviewers cannot determine whether an intended attribute was absent from the source population, removed structurally, excluded by policy, or simply outranked near the form boundary. Preserving those distinctions turns the evaluator into an object that can be compared and revised. It also allows a later representation or policy to be applied to the same source population without introducing stochastic regeneration as another source of change.

For computational psychometrics, population formation and the later evaluator should be treated as separable but interacting objects of investigation because each can change the construct evidence, content, and exact wording preserved, removed, or displaced before review. Inspecting the evaluator and its interconnected decisions should therefore be a first-class psychometric task, not merely a technical prelude to validation.

These studies establish dependence within the configured evaluator: they identify how source-population and evaluator choices alter the material available for judgment, without treating that material as evidence of a finished instrument's quality or validity.

The next empirical stage is evaluation by psychometricians and respondents. Those studies can test whether the semantic contrasts, structural trajectories, coverage patterns, and candidate-form differences observed here anticipate item functioning, dimensional structure, response processes, or expert judgments. Their findings can then motivate targeted revisions to the items or to the evaluator itself. Reapplying the evaluator after each revision would reveal how those changes alter the evidence, content coverage, and exact items presented in the next review cycle. This creates a loop of investigation linking empirical instrument development to the computational decisions that shape each candidate form. The present studies provide the traceability needed to make that loop inspectable.

\hypertarget{limitations-and-future-directions}{%
\section{Limitations and Future Directions}\label{limitations-and-future-directions}}

The evidence comes from one Big Five construct framework, two generators, Qwen3.5-27B and Gemma 3 27B-IT, two generation packages, five embedding configurations, two structural methods, two candidate-form eligibility policies, and one candidate-form procedure. The 20 attributes made the evaluator's decisions inspectable, but they do not exhaust the possible structure or content of the Big Five. Other constructs, languages, item formats, model families, anchor systems, pool sizes, structural methods, and review constraints may produce different patterns, and the observed configuration dependence gives reason to expect that variation rather than treating it as residual noise.

The evaluator-configuration comparisons do not isolate individual model properties. The generators differ in training, architecture, tokenizer, post-training, and other properties, so their contrast characterizes realized source populations rather than a general model ranking. Embedding family, training, architecture, parameter count, dimension, input formatting, and geometry also differ in bundles. The same-dimension comparison shows that vector dimension alone is insufficient to explain the observed differences, but it does not identify which remaining property does. The two generation packages likewise combine prompt content, supplied construct information, adaptive instructions, and different generation gates. The results should be interpreted as dependence on complete evaluator configurations, which makes isolating consequential components a priority for further investigation.

All evaluator evidence was derived before respondent-data validation. Semantic alignment, network organization, retained coverage, stability, and candidate-form composition do not establish item functioning, score reliability, dimensional structure in respondents, measurement invariance, or validity for a proposed use. Those claims require review by psychometricians, response-process evidence, and appropriately designed respondent studies. The present contribution concerns the evidence and material made available for those later stages; a natural extension is to follow candidate items through psychometric review and, eventually, respondent studies.

The resampling procedures characterize variation across repeated generation tasks, not uncertainty from participant sampling. The single BGE-M3/TMFG structural analytic noncompletion remained non-evaluable wherever paired evidence was required, and its partial trace was not converted into removal, method disagreement, or form evidence. Although this treatment limits complete comparison in the affected configuration, it preserves the distinction between negative evidence and absent evidence; the noncompletion is also a practical finding because computational evaluators can fail analytically in practice, and an operational workflow must preserve unknown outcomes rather than convert them into rejection.

Finally, the complete candidate forms depended on large source populations and a quota of four items per content cell. Smaller pools, more restrictive content rules, different quotas, or additional wording constraints may create underfilled cells and alter the relative consequences of inclusive and agreement policies. The bounded source-population calibration supports the present design as a pragmatic workflow choice, not as a general optimum. Future work should vary review capacity and pool depth while retaining task-specific item identity, because the current results suggest that narrow cutoffs can magnify otherwise modest local differences.

\hypertarget{open-science-data-and-materials}{%
\section{Open Science, Data, and Materials}\label{open-science-data-and-materials}}

The analysis-ready data and Python analysis code supporting the results reported in this article are publicly available through Zenodo at \href{https://doi.org/10.5281/zenodo.21968239}{10.5281/zenodo.21968239}. The deposit includes exact generation prompts and model responses, parsed candidate and selection records, embeddings, structural-evaluation evidence, candidate-form outputs, a data dictionary, provenance documentation, and pinned software requirements.

\hypertarget{conclusion}{%
\section{Conclusion}\label{conclusion}}

AI-assisted item generation makes item production scalable while increasing the consequences of how candidate populations are formed and evaluated. Across two studies, we found that generators and generation packages shaped the source populations entering the evaluator, and that representation sensitivity differed across those populations. Representation then changed the construct evidence attached to fixed items and the content that survived structural reduction. Both candidate-form policies filled all content cells in all 19 evaluable configurations, but they presented different items for judgment. Changing the embedding configuration, which also changed the structural evidence and ranking derived from that representation, produced greater observed candidate-form divergence than changing the form policy. Together, these findings place both source-population formation and the configuration of the later evaluator within the measurement design.

The consequential question is therefore not simply whether computational screening should precede expert review. It is what that screening represents, what it removes, what it includes in a candidate form, and whether those decisions remain inspectable. When task-specific item identity, semantic contrasts, structural trajectories, policy decisions, displaced content, and unknown evidence are preserved, the evaluator can be studied and improved as a psychometric object. Psychometricians then receive not only a manageable set of items, but also a clearer account of how that set came to exist.

\section*{Acknowledgments}

This research was supported in part through computational resources and services provided by Advanced Research Computing at the University of Michigan, Ann Arbor. Advanced Research Computing RRID: \href{https://documentation.its.umich.edu/node/4980}{SCR\_027337}.

\clearpage
\bibliographystyle{mslapa}
\bibliography{references}

\clearpage
\appendix
\makeatletter
\@addtoreset{table}{section}
\makeatother
\renewcommand{\thesection}{Appendix \Alph{section}.}
\renewcommand{\thesubsection}{\Alph{section}.\arabic{subsection}}
\renewcommand{\thesubsubsection}{\thesubsection.\arabic{subsubsection}}
\renewcommand{\thetable}{\Alph{section}\arabic{table}}
\renewcommand{\theHtable}{appendix.\Alph{section}.\arabic{table}}
\hypertarget{appendix-a.-construct-framework-and-generation-instruments}{%
\section{Construct Framework and Generation Instruments}\label{appendix-a.-construct-framework-and-generation-instruments}}

This appendix describes the construct framework and generation packages needed to interpret the two studies. It distinguishes the attributes used to organize the evaluator from established Big Five taxonomies and clarifies how the two packages differed.

\hypertarget{construct-framework}{%
\subsection{Construct Framework}\label{construct-framework}}

The study uses the five Big Five traits and four attributes within each trait. These 20 attributes organize generation quotas, semantic anchors, retained-content summaries, and candidate-form cells. They are working targets for this evaluator rather than a proposed facet taxonomy.

The intended response process is an adult rating a self-descriptive personality item on a Likert agreement scale. Item wording was expected to describe a disposition in first-person language without response options or list structure. Table~\ref{tab:A1} gives the trait definitions and principal boundaries, and Table~\ref{tab:A2} documents the attribute definitions, behavioral indicators, and principal confounds that delimit the working targets. Attribute definitions and behavioral indicators supplied the construct-indirect generation guidance and the declared semantic comparisons; the boundaries and confounds clarify the intended content domain.

\Needspace{10\baselineskip}
\begingroup
\setlength{\tabcolsep}{2.00pt}
\renewcommand{\arraystretch}{1.08}
\begin{longtable}[]{@{}
  >{\raggedright\arraybackslash\hyphenpenalty=10000\exhyphenpenalty=10000}p{(\columnwidth - 4\tabcolsep) * \real{0.2000}}
  >{\raggedright\arraybackslash}p{(\columnwidth - 4\tabcolsep) * \real{0.4000}}
  >{\raggedright\arraybackslash}p{(\columnwidth - 4\tabcolsep) * \real{0.4000}}@{}}
\caption{Trait definitions and boundaries}\label{tab:A1} \\
\toprule\noalign{}
\begin{minipage}[b]{\linewidth}\raggedright
Trait
\end{minipage} & \begin{minipage}[b]{\linewidth}\raggedright
Definition
\end{minipage} & \begin{minipage}[b]{\linewidth}\raggedright
Principal boundaries
\end{minipage} \\
\midrule\noalign{}
\endfirsthead
\toprule\noalign{}
\begin{minipage}[b]{\linewidth}\raggedright
Trait
\end{minipage} & \begin{minipage}[b]{\linewidth}\raggedright
Definition
\end{minipage} & \begin{minipage}[b]{\linewidth}\raggedright
Principal boundaries
\end{minipage} \\
\midrule\noalign{}
\endhead
\bottomrule\noalign{}
\endlastfoot
Agreeableness & Tendency toward cooperation, compassion, trust, humility, and concern for interpersonal harmony & Not compliance alone, conflict avoidance alone, or lack of assertiveness \\
\mbox{Conscientiousness} & Tendency toward organization, dependability, self-control, and goal-directed behavior & Not workaholism, social conformity alone, or socioeconomic status \\
Extraversion & Tendency toward sociability, positive affect, assertive social engagement, and energetic activity & Not popularity, a leadership title, or social skill alone \\
Neuroticism & Tendency toward negative emotional reactivity, worry, insecurity, and emotional volatility & Not a clinical diagnosis, temporary stress alone, or trauma exposure \\
Openness & Tendency toward imagination, intellectual curiosity, aesthetic sensitivity, and interest in ideas or experiences & Not general intelligence, education level, or political ideology \\
\end{longtable}
\endgroup

\Needspace{10\baselineskip}
\begingroup
\setlength{\tabcolsep}{1.50pt}
\renewcommand{\arraystretch}{1.08}
\begin{longtable}[]{@{}
  >{\raggedright\arraybackslash\hyphenpenalty=10000\exhyphenpenalty=10000}p{(\columnwidth - 8\tabcolsep) * \real{0.1550}}
  >{\raggedright\arraybackslash\hyphenpenalty=10000\exhyphenpenalty=10000}p{(\columnwidth - 8\tabcolsep) * \real{0.1400}}
  >{\raggedright\arraybackslash}p{(\columnwidth - 8\tabcolsep) * \real{0.2350}}
  >{\raggedright\arraybackslash}p{(\columnwidth - 8\tabcolsep) * \real{0.2900}}
  >{\raggedright\arraybackslash}p{(\columnwidth - 8\tabcolsep) * \real{0.1800}}@{}}
\caption{Attributes, behavioral indicators, and principal confounds}\label{tab:A2} \\
\toprule\noalign{}
\begin{minipage}[b]{\linewidth}\raggedright
Trait
\end{minipage} & \begin{minipage}[b]{\linewidth}\raggedright
Attribute
\end{minipage} & \begin{minipage}[b]{\linewidth}\raggedright
Definition
\end{minipage} & \begin{minipage}[b]{\linewidth}\raggedright
Behavioral indicators
\end{minipage} & \begin{minipage}[b]{\linewidth}\raggedright
Principal confounds
\end{minipage} \\
\midrule\noalign{}
\endfirsthead
\toprule\noalign{}
\begin{minipage}[b]{\linewidth}\raggedright
Trait
\end{minipage} & \begin{minipage}[b]{\linewidth}\raggedright
Attribute
\end{minipage} & \begin{minipage}[b]{\linewidth}\raggedright
Definition
\end{minipage} & \begin{minipage}[b]{\linewidth}\raggedright
Behavioral indicators
\end{minipage} & \begin{minipage}[b]{\linewidth}\raggedright
Principal confounds
\end{minipage} \\
\midrule\noalign{}
\endhead
\bottomrule\noalign{}
\endlastfoot
Agreeableness & \mbox{compassionate} & Concern for others' feelings, needs, and difficulties & notices when others struggle; responds to others' distress; considers another person's feelings & caregiving occupation; emotional contagion \\
Agreeableness & cooperative & Willingness to work with others and support shared goals & looks for common ground; helps group plans work; adjusts when collaboration requires it & obedience; low autonomy \\
Agreeableness & humble & Modest self-presentation and willingness to recognize others' contributions & credits others; admits mistakes; does not need to be the center of praise & low self-esteem; social anxiety \\
Agreeableness & trustworthy & Honest and dependable conduct in relationships & keeps confidences; keeps promises; acts honestly when no one is watching & reputation; legal compliance \\
\mbox{Conscientiousness} & disciplined & Ability to persist, regulate impulses, and continue tasks despite distractions & stays with difficult tasks; resists distractions; keeps routines & physical endurance; external supervision \\
\mbox{Conscientiousness} & organized & Preference for order, planning, and keeping tasks or materials arranged & keeps track of tasks; plans before acting; maintains order & living situation; administrative job demands \\
\mbox{Conscientiousness} & prudent & Careful consideration of consequences before acting & checks details before deciding; thinks through risks; avoids impulsive choices & risk aversion due to resources; anxiety \\
\mbox{Conscientiousness} & responsible & Reliability in meeting obligations and following through on commitments & follows through; meets commitments; takes obligations seriously & caretaking role; job seniority \\
Extraversion & assertive & Willingness to speak up, take initiative, and express preferences in groups & states preferences clearly; takes the floor when needed; offers direction in groups & dominance; aggression \\
Extraversion & energetic & Active, vigorous, and lively engagement with tasks or social situations & keeps momentum in activities; seeks active settings; maintains lively participation & physical fitness; sleep quality \\
Extraversion & friendly & Warm and approachable orientation toward social contact & starts pleasant interactions; makes others feel welcome; greets people easily & politeness norms; customer service role \\
Extraversion & positive & Tendency to experience and express upbeat affect & finds enjoyment in ordinary events; brings enthusiasm to activities; looks forward to plans & optimism bias; mania \\
Neuroticism & anxious & Tendency to worry, anticipate problems, or feel tense & worries before events; expects things to go wrong; feels tense under uncertainty & clinical anxiety disorder; realistic danger \\
Neuroticism & depressed & Tendency toward sadness, discouragement, or low mood in everyday life & feels discouraged; has trouble shaking off low mood; dwells on disappointments & clinical depression diagnosis; bereavement \\
Neuroticism & emotional & Tendency toward strong, fluctuating, or easily triggered emotional reactions & feelings shift quickly; small events affect mood; reacts strongly to setbacks & empathy; expressiveness alone \\
Neuroticism & insecure & Tendency to doubt oneself or feel vulnerable to criticism and rejection & second-guesses self; feels shaken by criticism; questions whether others accept them & social status; specific skill gaps \\
Openness & creative & Preference for originality, invention, and generating novel approaches & tries unusual approaches; imagines alternatives; finds original uses or solutions & artistic skill; occupational creativity \\
Openness & curious & Interest in learning, questioning, and seeking information & seeks explanations; asks follow-up questions; explores unfamiliar topics & school achievement; general knowledge \\
Openness & perceptual & Attention to subtle patterns, nuances, aesthetics, or sensory details & notices subtle differences; attends to patterns; takes in sensory details & sensory acuity; visual ability \\
Openness & philosophical & Interest in abstract questions about meaning, values, existence, and ideas & reflects on big questions; considers abstract possibilities; thinks about values and meaning & religious identity; academic philosophy training \\
\end{longtable}
\endgroup

\hypertarget{anchor-families}{%
\subsection{Anchor Families}\label{anchor-families}}

The semantic anchor panel was derived from the construct framework before item outcomes were examined. It contained the five trait labels and definitions, the 20 attribute labels and definitions, and one behavioral-indicator summary for each attribute. Each anchor was encoded separately within each embedding configuration.

For Study 1, the five anchor families remained separate so that direct lexical cues could be distinguished from richer construct descriptions. Within a family, the intended target was contrasted with the strongest alternative in the comparison set. Study 2 ranked candidate items using equal contributions from the attribute-definition and behavioral-indicator contrasts. This design made the comparison set explicit and preserved the distinction between similarity to a target and evidence that the target exceeded plausible alternatives.

\hypertarget{generation-packages}{%
\subsection{Generation Packages}\label{generation-packages}}

The two generation packages shared the same generators, adaptive request structure, target quotas, output parsing, and basic item constraints. They differed in the construct information supplied and in the treatment of direct construct terminology.

The AI-GENIE source package followed the source workflow's compact construct naming and adaptive anti-repetition structure. It supplied the trait and attribute labels and permitted those labels to appear in candidate items. Earlier eligible items were returned to later requests to encourage novelty.

The construct-indirect package supplied the attribute definitions and behavioral indicators shown in Table~\ref{tab:A2}, together with additional item-writing guidance. Its central instruction asked the model to express each target through concrete behaviors, preferences, reactions, or experiences instead of naming the construct. The package's generation gate separately prohibited the declared trait and attribute terms in respondent-facing items. The comparison therefore concerns two complete generation packages, not an isolated prompt manipulation.

The complete human-readable templates are reproduced below. Dollar-sign names identify values inserted for a particular request. The templates retain \passthrough{\lstinline!statement!} as the required JSON field name; the article calls the resulting content items.

\hypertarget{ai-genie-source-template}{%
\subsubsection{AI-GENIE Source Template}\label{ai-genie-source-template}}

System instruction:

\begin{lstlisting}
You are an expert measurement methodologist; more specifically, you are an accomplished, well-trained, and knowledgeable scale-developer specializing in personality assessment. Your task is to create novel, high-quality, and robust items for a new inventory called 'Big Five Personality Inventory.' Ensure the items are appropriate for an audience of adults responding to a self-report personality questionnaire.
For your reference, the response options for the items you are authoring will be as follows: (1) Strongly Disagree, (2) Disagree, (3) Neither Disagree nor Agree, (4) Agree, and (5) Strongly Agree. I will add the response options myself AFTER you create the items; do NOT include them in the items you write. However, you should still ensure the items are appropriately phrased given these options.
\end{lstlisting}

Initial user instruction:

\begin{lstlisting}
Generate a grand total of $items_total novel, UNIQUE, reliable, and valid personality items for a scale called 'Big Five Personality Inventory.' This inventory will be administered to an audience of adults. Write items related to the attributes of the item type '$trait_label'. Here are the attributes of the item type '$trait_label': $numbered_attribute_list. Generate EXACTLY $items_per_attribute_display items PER attribute. Use the $attribute_count attributes EXACTLY as provided; do NOT add your own or leave any out.
EACH item should be ROBUST, NOVEL, and UNIQUE. These items must be top-quality.
Ensure that each item is extremely high-quality, psychometrically robust, and concise. Each item should be novel, so be creative; aim for BREADTH across these attributes, deliberately varying wording, sentence structure, and the facet of the attribute each item emphasizes. Avoid producing items that are minor rephrasings of one another. Items should be polished and ready for immediate practical use.
Return output STRICTLY as a JSON array of objects, each with keys `attribute` and `statement`, e.g.:
[{"attribute":"$first_attribute","statement":"Your item here."}, ...]
This JSON formatting is EXTREMELY important. ONLY output the items in this formatting; DO NOT include any other text in your response. The "attribute" key should ONLY have these EXACT values: $attribute_csv.
\end{lstlisting}

Continuation appended to later user requests:

\begin{lstlisting}
Do NOT repeat, rephrase, or reuse the content of ANY items from this list of items you've already generated for $trait_id:
$existing_items
\end{lstlisting}

\hypertarget{construct-indirect-template}{%
\subsubsection{Construct-Indirect Template}\label{construct-indirect-template}}

System instruction:

\begin{lstlisting}
You are an expert psychometrician and scale developer. Write survey items that express construct-relevant behaviors without giving away the intended label through obvious wording.
\end{lstlisting}

User instruction:

\begin{lstlisting}
Generate exactly $items_total Big Five personality items for the trait "$trait_label".

Use these attributes as the hidden item targets and JSON labels: $attribute_csv.
Generate exactly $items_per_attribute items for each attribute.

Construct-indirect wording:
- The JSON "attribute" value must use the exact target label, but the item "statement" must not use the trait name, trait label, attribute label, or close obvious derivatives.
- Do not use these terms in item statements: $prohibited_terms_csv.
- Express each target through concrete behaviors, preferences, reactions, or experiences instead of naming the construct.

Construct specification guidance:
$construct_guidance

General item-writing requirements:
$configured_requirements
- Each item must express one behavior, belief, preference, feeling, or tendency.
- Each item must target exactly one listed attribute.
- Do not combine multiple attributes in a single item.
- Avoid absolute terms such as "always" and "never" unless they are necessary.
- Write items likely to show response variability among adults.
- Preserve breadth within each attribute by varying situations, wording, and behavioral manifestations.
- Avoid direct duplicates and close paraphrases of existing items.

Existing items to avoid:
$existing_items

Return output STRICTLY as a JSON array of objects. Each object must contain exactly:
- "attribute": one of $attribute_csv
- "statement": the item text

Example:
[{"attribute":"$first_attribute","statement":"Your item here."}]

Do not include prose, markdown, numbering, comments, or any text outside the JSON array.
\end{lstlisting}

\Needspace{10\baselineskip}
\begin{longtable}[]{@{}
  >{\raggedright\arraybackslash}p{(\columnwidth - 2\tabcolsep) * \real{0.5000}}
  >{\raggedright\arraybackslash}p{(\columnwidth - 2\tabcolsep) * \real{0.5000}}@{}}
\caption{Prompt-template variables}\label{tab:A3} \\
\toprule\noalign{}
\begin{minipage}[b]{\linewidth}\raggedright
Variable
\end{minipage} & \begin{minipage}[b]{\linewidth}\raggedright
Inserted content
\end{minipage} \\
\midrule\noalign{}
\endfirsthead
\toprule\noalign{}
\begin{minipage}[b]{\linewidth}\raggedright
Variable
\end{minipage} & \begin{minipage}[b]{\linewidth}\raggedright
Inserted content
\end{minipage} \\
\midrule\noalign{}
\endhead
\bottomrule\noalign{}
\endlastfoot
\passthrough{\lstinline!$items\_total!} & Number of items requested in one generation response \\
\passthrough{\lstinline!$trait\_label!}, \passthrough{\lstinline!$trait\_id!} & Display label and identifier for the target trait \\
\passthrough{\lstinline!$numbered\_attribute\_list!}, \passthrough{\lstinline!$attribute\_count!}, \passthrough{\lstinline!$attribute\_csv!}, \passthrough{\lstinline!$first\_attribute!} & The four attributes for the target trait, rendered as required by each part of the prompt \\
\passthrough{\lstinline!$items\_per\_attribute\_display!}, \passthrough{\lstinline!$items\_per\_attribute!} & Number of items requested for each attribute \\
\passthrough{\lstinline!$construct\_guidance!} & Attribute definitions and behavioral indicators for the target trait \\
\passthrough{\lstinline!$configured\_requirements!} & Applicable item-length, self-report, formatting, double-barreled, and evaluative-wording instructions \\
\passthrough{\lstinline!$prohibited\_terms\_csv!} & Trait, attribute, obvious derivative, and additional prohibited terms for the construct-indirect package \\
\passthrough{\lstinline!$existing\_items!} & Earlier items from the same adaptive generation task \\
\end{longtable}

\hypertarget{adaptive-histories-and-quotas}{%
\subsection{Adaptive Histories and Quotas}\label{adaptive-histories-and-quotas}}

A generation task was an adaptive sequence defined by generator, generation package, trait, and repetition. Later requests received recent items from the eligible item pool for the same task, so each response affected the context of subsequent generation. This shared history is why the generation task, rather than the individual item, served as the repeated unit in the bootstrap and resampling analyses.

Each request sought two candidate items for each of the trait's four attributes. A task continued until it filled a quota of 20 selected items per attribute, yielding 80 items in its source population. Eligible items arriving after an attribute quota was full remained in the eligible item pool as surplus rather than being reclassified as rejected. Across the two generators, two generation packages, five traits, and 20 repetitions, the design produced 400 generation tasks and 32,000 selected item occurrences and 1,226 eligible surplus item occurrences.

\hypertarget{eligibility-rules-and-review-warnings}{%
\subsection{Eligibility Rules and Review Warnings}\label{eligibility-rules-and-review-warnings}}

The generation gate determined whether a candidate item could enter the eligible item pool. An item had to be parseable, assigned to one of the declared attributes, nonempty, between 4 and 24 words, and free of response-option or numbered-list artifacts. Exact normalized duplicates were excluded within the applicable adaptive history. Under the construct-indirect package, declared trait and attribute terms were also prohibited. Quota exhaustion was recorded separately from rejection so that an eligible item did not become a rejected item merely because its content cell was already full.

Review warnings preserved information for judgment by psychometricians without excluding an item. These warnings identified weakened self-report form, possible double-barreled content, negation or reverse wording, and strongly evaluative language. They accompanied eligible items into later ranking and reporting but were not converted into an unvalidated numerical measure of item quality.

The generation gate and review warnings make different claims. A gate rule verifies a formal or lexical condition that it explicitly encodes. A warning directs a psychometrician's attention to a possible wording concern. Neither determines the adequacy of the item, and neither substitutes for the contrastive or structural evidence examined in the two studies.

\clearpage
\hypertarget{appendix-b.-evaluator-configuration-and-candidate-form-construction}{%
\section{Evaluator Configuration and Candidate-Form Construction}\label{appendix-b.-evaluator-configuration-and-candidate-form-construction}}

This appendix describes the evaluator choices that materially affect the comparisons: the configured model panel, the structural procedures, the treatment of incomplete evidence, the two eligibility policies, the anchor-based ranking rule, the joint assignment procedure, and the generation-task resampling analysis.

\hypertarget{configured-model-panel}{%
\subsection{Configured Model Panel}\label{configured-model-panel}}

The configuration question is which representation systems entered the evaluator and what scientific role each played. Qwen3.5-27B and Gemma 3 27B-IT generated separate source populations, which were not pooled into a common candidate form. Every selected item was represented independently within each of five embedding configurations. The table's unit is one configured embedding system; it defines the design rather than summarizing item outcomes, and it cannot isolate the effect of any one bundled system property.

\Needspace{10\baselineskip}
\begingroup
\setlength{\tabcolsep}{2.50pt}
\renewcommand{\arraystretch}{1.08}
\begin{longtable}[]{@{}
  >{\raggedright\arraybackslash}p{(\columnwidth - 6\tabcolsep) * \real{0.2500}}
  >{\centering\arraybackslash}p{(\columnwidth - 6\tabcolsep) * \real{0.1800}}
  >{\raggedright\arraybackslash}p{(\columnwidth - 6\tabcolsep) * \real{0.2400}}
  >{\raggedright\arraybackslash}p{(\columnwidth - 6\tabcolsep) * \real{0.3300}}@{}}
\caption{Configured embedding panel and design roles.}\label{tab:B1} \\
\toprule\noalign{}
\begin{minipage}[b]{\linewidth}\raggedright
Embedding configuration
\end{minipage} & \begin{minipage}[b]{\linewidth}\raggedright
\mbox{Vector dimension}
\end{minipage} & \begin{minipage}[b]{\linewidth}\raggedright
Input mode
\end{minipage} & \begin{minipage}[b]{\linewidth}\raggedright
Design role
\end{minipage} \\
\midrule\noalign{}
\endfirsthead
\toprule\noalign{}
\begin{minipage}[b]{\linewidth}\raggedright
Embedding configuration
\end{minipage} & \begin{minipage}[b]{\linewidth}\raggedright
\mbox{Vector dimension}
\end{minipage} & \begin{minipage}[b]{\linewidth}\raggedright
Input mode
\end{minipage} & \begin{minipage}[b]{\linewidth}\raggedright
Design role
\end{minipage} \\
\midrule\noalign{}
\endhead
\bottomrule\noalign{}
\endlastfoot
Qwen3-Embedding-0.6B & 1,024 & Document & Small configured Qwen level \\
Qwen3-Embedding-4B & 2,560 & Document & Middle configured Qwen level \\
Qwen3-Embedding-8B & 4,096 & Document & Large configured Qwen level \\
EmbeddingGemma-300M & 768 & Model-specific document prompt & Compact comparison space \\
BGE-M3 & 1,024 & Document & Same-dimension comparison with Qwen 0.6B and an additional model family \\
\end{longtable}
\endgroup

The vector dimensions shown above, 768, 1,024, 2,560, and 4,096, were also supplied to EGA as its effective sample size. Graphical-lasso regularization therefore varied with the embedding configuration.

Model family, training, architecture, parameter count, dimension, input treatment, and other system properties remain bundled within these configurations. The Qwen models provide a bounded within-family comparison, while BGE-M3 and Qwen 0.6B provide a same-dimension diagnostic. Neither comparison isolates a component cause.

\hypertarget{outcome-definitions-and-units}{%
\subsection{Outcome Definitions and Units}\label{outcome-definitions-and-units}}

The measurement question is which property of the evaluator each recurring outcome describes and at what analytic unit. The table defines each unit and denominator locally because no single denominator applies across item-, task-, structural-evaluation-, and form-level outcomes. These definitions keep geometry, identity, community correspondence, and coverage distinct; they do not make one outcome a general measure of item quality.

\Needspace{10\baselineskip}
\begingroup
\setlength{\tabcolsep}{2.50pt}
\renewcommand{\arraystretch}{1.08}
\begin{longtable}[]{@{}
  >{\raggedright\arraybackslash}p{(\columnwidth - 2\tabcolsep) * \real{0.2500}}
  >{\raggedright\arraybackslash}p{(\columnwidth - 2\tabcolsep) * \real{0.7500}}@{}}
\caption{Outcome definitions and analytic units.}\label{tab:B2} \\
\toprule\noalign{}
\begin{minipage}[b]{\linewidth}\raggedright
Outcome
\end{minipage} & \begin{minipage}[b]{\linewidth}\raggedright
Definition and analytic unit
\end{minipage} \\
\midrule\noalign{}
\endfirsthead
\toprule\noalign{}
\begin{minipage}[b]{\linewidth}\raggedright
Outcome
\end{minipage} & \begin{minipage}[b]{\linewidth}\raggedright
Definition and analytic unit
\end{minipage} \\
\midrule\noalign{}
\endhead
\bottomrule\noalign{}
\endlastfoot
Target advantage & For one item and anchor family, cosine similarity to the intended anchor minus similarity to the strongest alternative in the comparison set. Positive values favor the intended target, zero is a tie, and negative values favor an alternative. Scaled values divide this margin by its sample standard deviation within the embedding configuration and anchor family without shifting zero. \\
Target-first status or rate & For one item, target-first status requires a strictly positive target advantage; a tie does not count. The rate is the proportion of items meeting that condition. \\
Target rank & For one item, one plus the number of alternatives more similar than the intended target. Rank 1 means that no alternative is closer, but can include a tie and is therefore distinct from target-first status. \\
Geometry agreement & For one generation task and pair of embedding configurations, the Spearman agreement between their orderings of all 3,160 pairwise similarities among the same 80 source items. \\
Retained-set overlap & For one generation task, structural method, and pair of embedding configurations, the proportion of distinct retained items shared by both configurations, summarized by Jaccard similarity. \\
Community correspondence & For one structural evaluation at a specified stage, agreement between the recovered communities and intended attribute assignments. Adjusted mutual information \shortcitewithlabel{AMI}{vinhEtAl2010ClusteringComparison} is the leading measure, with NMI and ARI as companion measures. Initial values use all 80 source items; final values use only retained items. \\
Content coverage & At a specified evaluator stage, whether and how many items represent each intended attribute. Structural summaries use one attribute within one structural evaluation; candidate-form summaries use one content cell within one form configuration. \\
\end{longtable}
\endgroup

\hypertarget{structural-evaluation}{%
\subsection{Structural Evaluation}\label{structural-evaluation}}

The structural question is how the two configured network methods supplied comparable evidence after a shared local-dependence step. TMFG and graphical lasso were applied separately to each generation task within each embedding configuration. The analytic unit is one generation task under one embedding configuration and one structural method. The common GENIE-compatible sequence first applied UVA to the same embedding-derived input within both method traces. Because UVA did not use the network-method setting, it yielded the same removals before TMFG and graphical lasso supplied separate post-UVA community and item-stability evidence and final community structures among retained items. The method-specific initial community structures for the full source pool served as pre-reduction baselines and were not inputs to UVA. Terminal retention could then diverge at the method-specific stability stage. The table records configured settings, not observed performance, and the settings do not make either method a truth criterion.

\Needspace{10\baselineskip}
\begingroup
\setlength{\tabcolsep}{2.50pt}
\renewcommand{\arraystretch}{1.08}
\begin{longtable}[]{@{}
  >{\raggedright\arraybackslash}p{(\columnwidth - 4\tabcolsep) * \real{0.2300}}
  >{\raggedright\arraybackslash}p{(\columnwidth - 4\tabcolsep) * \real{0.2700}}
  >{\raggedright\arraybackslash}p{(\columnwidth - 4\tabcolsep) * \real{0.5000}}@{}}
\caption{Structural-evaluation settings and scientific roles.}\label{tab:B3} \\
\toprule\noalign{}
\begin{minipage}[b]{\linewidth}\raggedright
Setting
\end{minipage} & \begin{minipage}[b]{\linewidth}\raggedright
Configured value
\end{minipage} & \begin{minipage}[b]{\linewidth}\raggedright
Scientific role
\end{minipage} \\
\midrule\noalign{}
\endfirsthead
\toprule\noalign{}
\begin{minipage}[b]{\linewidth}\raggedright
Setting
\end{minipage} & \begin{minipage}[b]{\linewidth}\raggedright
Configured value
\end{minipage} & \begin{minipage}[b]{\linewidth}\raggedright
Scientific role
\end{minipage} \\
\midrule\noalign{}
\endhead
\bottomrule\noalign{}
\endlastfoot
Structural methods & TMFG and graphical lasso & Alternative sparse-network representations of the same embedded population \\
EGA community algorithm & Walktrap \cite{ponsLatapy2006Walktrap} & Community estimation \\
Unidimensionality procedure & Louvain \shortcite{blondelEtAl2008FastUnfolding} & Configured dimensionality assessment \\
UVA cutoff & 0.20 & Local-dependence or redundancy threshold \\
BootEGA iterations & 100 & Stability resampling within a structural evaluation \\
Bootstrap seed & 123 & Fixed reproducibility constant \\
Item-stability cutoff & 0.75 & Stability-based reduction \\
Final stability & Calculated but non-gating & Diagnostic evidence after final estimation \\
\end{longtable}
\endgroup

For every item, the structural trace preserved whether it was removed through UVA, removed below the stability threshold, retained, or unknown. A result was accepted only when the evidence needed to interpret its terminal state was available. Missing execution, analytic noncompletion, and unknown item-level evidence remained distinct from removal. Final stability did not retroactively alter the retention decision used in candidate-form construction.

\hypertarget{paired-method-states-and-eligibility-policies}{%
\subsection{Paired Method States and Eligibility Policies}\label{paired-method-states-and-eligibility-policies}}

The policy question is how paired method outcomes become form eligibility without converting missing evidence into rejection. The analytic unit is one item occurrence with paired TMFG and graphical-lasso outcomes; only a complete method pair enters an eligible or ineligible denominator. Both focal policies required complete evidence from the two structural methods. The table defines evidentiary states and does not equate agreement with item quality.

\Needspace{10\baselineskip}
\begingroup
\setlength{\tabcolsep}{2.50pt}
\renewcommand{\arraystretch}{1.08}
\begin{longtable}[]{@{}
  >{\raggedright\arraybackslash}p{(\columnwidth - 6\tabcolsep) * \real{0.2200}}
  >{\raggedright\arraybackslash}p{(\columnwidth - 6\tabcolsep) * \real{0.2800}}
  >{\raggedright\arraybackslash}p{(\columnwidth - 6\tabcolsep) * \real{0.2500}}
  >{\raggedright\arraybackslash}p{(\columnwidth - 6\tabcolsep) * \real{0.2500}}@{}}
\caption{Paired structural-method states and candidate-form eligibility.}\label{tab:B4} \\
\toprule\noalign{}
\begin{minipage}[b]{\linewidth}\raggedright
TMFG outcome
\end{minipage} & \begin{minipage}[b]{\linewidth}\raggedright
Graphical-lasso outcome
\end{minipage} & \begin{minipage}[b]{\linewidth}\raggedright
Inclusive policy
\end{minipage} & \begin{minipage}[b]{\linewidth}\raggedright
Agreement policy
\end{minipage} \\
\midrule\noalign{}
\endfirsthead
\toprule\noalign{}
\begin{minipage}[b]{\linewidth}\raggedright
TMFG outcome
\end{minipage} & \begin{minipage}[b]{\linewidth}\raggedright
Graphical-lasso outcome
\end{minipage} & \begin{minipage}[b]{\linewidth}\raggedright
Inclusive policy
\end{minipage} & \begin{minipage}[b]{\linewidth}\raggedright
Agreement policy
\end{minipage} \\
\midrule\noalign{}
\endhead
\bottomrule\noalign{}
\endlastfoot
Retained & Retained & Eligible & Eligible \\
Retained & Removed & Eligible & Ineligible \\
Removed & Retained & Eligible & Ineligible \\
Removed & Removed & Ineligible & Ineligible \\
Unknown & Any outcome & Unknown & Unknown \\
Any outcome & Unknown & Unknown & Unknown \\
\end{longtable}
\endgroup

The inclusive policy made an item form-eligible when either method retained it. Support from both methods entered the ordering only when two items had exactly the same anchor score. The agreement policy made an item form-eligible only when both methods retained it. It used the same anchor evidence, warnings, wording-uniqueness constraint, joint assignment, and quotas as the inclusive policy, and it did not fill a shortfall with an item supported by only one method.

Secondary TMFG-only and graphical-lasso-only forms helped locate how the asymmetric retention patterns shaped the two focal policies. They did not define additional primary policies or establish a preferred structural method.

\hypertarget{anchor-based-ranking}{%
\subsection{Anchor-Based Ranking}\label{anchor-based-ranking}}

Ranking occurred within each content cell. Two anchor components entered the candidate score: the intended attribute definition and its behavioral-indicator summary. For each component, the item's similarity to the intended anchor was reduced by its similarity to the strongest alternative attribute within the same trait. A positive value placed the target ahead of every alternative in the comparison set, zero indicated a tie, and a negative value favored another attribute.

Each component was divided by the standard deviation of its raw target advantage across the complete 32,000-item canonical ordinary population for the corresponding embedding configuration and anchor estimand. The transformation was not mean-centered, preserving zero as a tie between the target and its strongest alternative. The candidate score gave equal weight to the scaled definition and behavioral-indicator advantages, and both components had to be finite and available.

Form-eligible items were ordered first by higher anchor score. Under the inclusive policy, retention by both methods preceded single-method support only at an exact score tie. Lower warning burden supplied the next ordering rule, followed by complete generation-task provenance only when the preceding objectives remained tied. Structural corroboration therefore did not override a nonzero difference in anchor evidence, and source order served only as a deterministic convention.

\hypertarget{joint-candidate-form-assignment}{%
\subsection{Joint Candidate-Form Assignment}\label{joint-candidate-form-assignment}}

Items retained task-specific identities, so the same normalized wording generated in different adaptive histories remained distinct in the analysis. Normalized wording was used separately to prevent the same text from occupying more than one form position. The 20 content cells were defined by crossing five traits with four attributes per trait; for example, Conscientiousness/disciplined formed one content cell.

Assignment was solved jointly across all cells because the same normalized wording could appear among multiple form-eligible items. Each position could receive at most one item, an item could fill only its assigned content cell, and a normalized wording could appear at most once in the complete form. Any unfilled position remained explicit.

The assignment first maximized the number of cells receiving their first primary item, followed by the second primary, first alternate, and second alternate. Among equally complete assignments, it minimized displacement from the within-cell ranking and used source provenance only for an otherwise exact tie. This ordering prevented an arbitrary sequence of cells from consuming a recurring wording when assigning it elsewhere would preserve more complete coverage.

Each cell sought two primary and two alternate items, yielding 40 primary and 40 alternate positions in a complete form. A form position is therefore a primary or alternate slot, not an item's rank among all candidates in a cell. These quotas define the review object examined in this study; they are not an empirically optimized scale length.

\hypertarget{generation-task-resampling}{%
\subsection{Generation-Task Resampling}\label{generation-task-resampling}}

Candidate-form stability was examined through 1,000 deterministic resamples of generation tasks. Tasks were sampled with replacement within generator-by-generation-package-by-trait strata. The same numbered draw was paired across the five embedding configurations, and both eligibility policies were rebuilt from the common draw.

When a task was sampled more than once, each resampled copy of its items received distinct analysis identities while retaining the original task provenance. The analysis preserved form completeness, exact form overlap, warning burden, and source concentration. If a sampled task contained missing structural evidence, the affected comparison remained non-evaluable. The resulting distributions describe dependence on the mixture of repeated generation tasks; they do not represent participant sampling.

\clearpage
\hypertarget{appendix-c.-study-1-supporting-evidence}{%
\section{Study 1 Supporting Evidence}\label{appendix-c.-study-1-supporting-evidence}}

This appendix provides the bounded evidence needed to interpret Study 1's linked claims about construct representation, retained content, and recovered organization. The displays separate evidence layers that use different analytic units. They describe the behavior of the configured computational evaluator before respondent data or expert adjudication and do not establish item quality, dimensional validity, or a generally preferred embedding or structural method.

\hypertarget{construct-evidence-across-representations-and-anchors}{%
\subsection{Construct Evidence Across Representations and Anchors}\label{construct-evidence-across-representations-and-anchors}}

The first scientific question is whether the five configured representations supplied the same evidence when an intended construct was compared with plausible alternatives, and whether that conclusion depended on how the construct was represented. Table~\ref{tab:C1} uses all 32,000 selected item occurrences from 400 generation tasks. Each cell is the mean target advantage after division by its sample standard deviation within the embedding configuration and anchor family, without moving zero; positive values favor the intended target over its strongest competitor. The unit is an item-by-anchor evaluation, summarized across the fixed generation design. These contrasts assess alignment with the configured anchors, not respondent interpretation or psychometric validity, and the embedding configurations differ in several bundled properties.

\Needspace{10\baselineskip}
\begingroup
\setlength{\tabcolsep}{2.00pt}
\renewcommand{\arraystretch}{1.05}
\begin{longtable}[]{@{}
  >{\raggedright\arraybackslash}p{(\columnwidth - 10\tabcolsep) * \real{0.2500}}
  >{\raggedleft\arraybackslash}p{(\columnwidth - 10\tabcolsep) * \real{0.1500}}
  >{\raggedleft\arraybackslash}p{(\columnwidth - 10\tabcolsep) * \real{0.1500}}
  >{\raggedleft\arraybackslash}p{(\columnwidth - 10\tabcolsep) * \real{0.1500}}
  >{\raggedleft\arraybackslash}p{(\columnwidth - 10\tabcolsep) * \real{0.1500}}
  >{\raggedleft\arraybackslash}p{(\columnwidth - 10\tabcolsep) * \real{0.1500}}@{}}
\caption{Scaled target advantage across the five embedding configurations and five anchor families.}\label{tab:C1} \\
\toprule\noalign{}
\begin{minipage}[b]{\linewidth}\raggedright
Embedding configuration
\end{minipage} & \begin{minipage}[b]{\linewidth}\raggedleft
Trait label
\end{minipage} & \begin{minipage}[b]{\linewidth}\raggedleft
Trait definition
\end{minipage} & \begin{minipage}[b]{\linewidth}\raggedleft
Attribute label
\end{minipage} & \begin{minipage}[b]{\linewidth}\raggedleft
Attribute definition
\end{minipage} & \begin{minipage}[b]{\linewidth}\raggedleft
Behavioral indicator
\end{minipage} \\
\midrule\noalign{}
\endfirsthead
\toprule\noalign{}
\begin{minipage}[b]{\linewidth}\raggedright
Embedding configuration
\end{minipage} & \begin{minipage}[b]{\linewidth}\raggedleft
Trait label
\end{minipage} & \begin{minipage}[b]{\linewidth}\raggedleft
Trait definition
\end{minipage} & \begin{minipage}[b]{\linewidth}\raggedleft
Attribute label
\end{minipage} & \begin{minipage}[b]{\linewidth}\raggedleft
Attribute definition
\end{minipage} & \begin{minipage}[b]{\linewidth}\raggedleft
Behavioral indicator
\end{minipage} \\
\midrule\noalign{}
\endhead
\bottomrule\noalign{}
\endlastfoot
Qwen3-Embedding-0.6B & -0.209 & 0.349 & 0.656 & 0.914 & 0.887 \\
Qwen3-Embedding-4B & 0.129 & 0.508 & 0.828 & 1.037 & 1.408 \\
Qwen3-Embedding-8B & 0.309 & 0.639 & 1.014 & 1.076 & 1.417 \\
EmbeddingGemma-300M & -0.178 & 0.355 & 0.572 & 0.869 & 1.063 \\
BGE-M3 & -0.667 & 0.319 & 0.173 & 0.549 & 0.939 \\
\end{longtable}
\endgroup

The representation and anchor family changed the evidence together. Qwen 8B had the largest mean target advantage in each column, but the difference between literal labels and richer construct descriptions was also substantial. Three configurations produced negative mean target advantage for trait labels while remaining positive for trait definitions. The result supports competitor-based comparisons using construct representations that match the intended content; it does not support treating an isolated target cosine as construct evidence.

An item-to-item comparison provided a related anchor-independent check on the source populations. Within each task and embedding, construct-indirect items showed greater average similarity within their assigned attribute than between different assigned attributes in all ten generator-system-by-embedding comparisons. The construct-indirect advantage ranged from .008 to .032 cosine units within the relevant embedding spaces. This result shows that the compound package changed the realized semantic organization of the source population; it does not establish that the later network procedure would recover fewer communities, retain more items, or produce a more valid instrument.

\Needspace{14\baselineskip}
\begingroup
\setlength{\tabcolsep}{2.00pt}
\renewcommand{\arraystretch}{1.05}
\begin{longtable}[]{@{}
  >{\raggedright\arraybackslash}p{(\columnwidth - 8\tabcolsep) * \real{0.1000}}
  >{\raggedright\arraybackslash}p{(\columnwidth - 8\tabcolsep) * \real{0.1800}}
  >{\raggedleft\arraybackslash}p{(\columnwidth - 8\tabcolsep) * \real{0.1600}}
  >{\raggedleft\arraybackslash}p{(\columnwidth - 8\tabcolsep) * \real{0.1600}}
  >{\raggedleft\arraybackslash}p{(\columnwidth - 8\tabcolsep) * \real{0.4000}}@{}}
\caption{Within-attribute minus between-attribute item similarity by generator, generation package, and embedding configuration. Positive values indicate that items assigned to the same attribute were more similar to one another than to items assigned to other attributes within the same trait. Brackets contain equal-tailed 95\% task-bootstrap intervals from the existing compound-package contrasts. Cosine levels are interpreted only within an embedding configuration.}\label{tab:C2} \\
\toprule\noalign{}
\begin{minipage}[b]{\linewidth}\raggedright
Generator
\end{minipage} & \begin{minipage}[b]{\linewidth}\raggedright
Embedding
\end{minipage} & \begin{minipage}[b]{\linewidth}\raggedleft
\mbox{AI-GENIE source}
\end{minipage} & \begin{minipage}[b]{\linewidth}\raggedleft
\mbox{Construct indirect}
\end{minipage} & \begin{minipage}[b]{\linewidth}\raggedleft
Indirect - source {[}95\% interval{]}
\end{minipage} \\
\midrule\noalign{}
\endfirsthead
\toprule\noalign{}
\begin{minipage}[b]{\linewidth}\raggedright
Generator
\end{minipage} & \begin{minipage}[b]{\linewidth}\raggedright
Embedding
\end{minipage} & \begin{minipage}[b]{\linewidth}\raggedleft
\mbox{AI-GENIE source}
\end{minipage} & \begin{minipage}[b]{\linewidth}\raggedleft
\mbox{Construct indirect}
\end{minipage} & \begin{minipage}[b]{\linewidth}\raggedleft
Indirect - source {[}95\% interval{]}
\end{minipage} \\
\midrule\noalign{}
\endhead
\bottomrule\noalign{}
\endlastfoot
Qwen & BGE-M3 & .069 & .101 & .032 {[}.029, .034{]} \\
Qwen & Qwen 0.6B & .112 & .132 & .020 {[}.017, .023{]} \\
Qwen & Qwen 4B & .113 & .133 & .020 {[}.017, .023{]} \\
Qwen & Qwen 8B & .128 & .152 & .024 {[}.021, .028{]} \\
Qwen & EmbeddingGemma & .106 & .137 & .031 {[}.028, .034{]} \\
Gemma & BGE-M3 & .066 & .081 & .015 {[}.013, .016{]} \\
Gemma & Qwen 0.6B & .108 & .116 & .008 {[}.006, .010{]} \\
Gemma & Qwen 4B & .106 & .115 & .008 {[}.006, .010{]} \\
Gemma & Qwen 8B & .121 & .133 & .012 {[}.009, .014{]} \\
Gemma & EmbeddingGemma & .101 & .112 & .011 {[}.009, .013{]} \\
\end{longtable}
\endgroup

\hypertarget{behavioral-indicator-construct-evidence}{%
\subsubsection{Behavioral-Indicator Construct Evidence}\label{behavioral-indicator-construct-evidence}}

Behavioral indicators provide a compact view of three construct-evidence measures because Qwen 4B and Qwen 8B were nearly tied in average target advantage. Table~\ref{tab:C3} asks whether that near tie also appeared in strict target-first status and target rank. The unit is one of the same 32,000 items evaluated against its intended behavioral indicator and the other three attributes within its trait; the denominator is 32,000 items from 400 observed generation tasks, with no missing construct evidence. Target-first status requires a strictly positive advantage, whereas rank 1 can include a tie. These are anchor-relative screening results rather than estimates of respondent understanding or item validity.

\Needspace{10\baselineskip}
\begingroup
\setlength{\tabcolsep}{2.00pt}
\renewcommand{\arraystretch}{1.05}
\begin{longtable}[]{@{}
  >{\raggedright\arraybackslash}p{(\columnwidth - 6\tabcolsep) * \real{0.3100}}
  >{\raggedleft\arraybackslash}p{(\columnwidth - 6\tabcolsep) * \real{0.2300}}
  >{\raggedleft\arraybackslash}p{(\columnwidth - 6\tabcolsep) * \real{0.2300}}
  >{\raggedleft\arraybackslash}p{(\columnwidth - 6\tabcolsep) * \real{0.2300}}@{}}
\caption{Behavioral-indicator construct evidence across embedding configurations.}\label{tab:C3} \\
\toprule\noalign{}
\begin{minipage}[b]{\linewidth}\raggedright
Embedding configuration
\end{minipage} & \begin{minipage}[b]{\linewidth}\raggedleft
Scaled target advantage
\end{minipage} & \begin{minipage}[b]{\linewidth}\raggedleft
Target-first rate
\end{minipage} & \begin{minipage}[b]{\linewidth}\raggedleft
Mean target rank
\end{minipage} \\
\midrule\noalign{}
\endfirsthead
\toprule\noalign{}
\begin{minipage}[b]{\linewidth}\raggedright
Embedding configuration
\end{minipage} & \begin{minipage}[b]{\linewidth}\raggedleft
Scaled target advantage
\end{minipage} & \begin{minipage}[b]{\linewidth}\raggedleft
Target-first rate
\end{minipage} & \begin{minipage}[b]{\linewidth}\raggedleft
Mean target rank
\end{minipage} \\
\midrule\noalign{}
\endhead
\bottomrule\noalign{}
\endlastfoot
Qwen3-Embedding-0.6B & .887 & 81.00\% & 1.273 \\
Qwen3-Embedding-4B & 1.408 & 92.52\% & 1.096 \\
Qwen3-Embedding-8B & 1.417 & 92.16\% & 1.095 \\
EmbeddingGemma-300M & 1.063 & 85.64\% & 1.199 \\
BGE-M3 & .939 & 83.05\% & 1.235 \\
\end{longtable}
\endgroup

Across the five embedding configurations, target-first status varied for 5,639 of 16,000 AI-GENIE-source items (35.2\%) and 4,078 of 16,000 construct-indirect items (25.5\%). The fixed wording and target assignments therefore did not guarantee the same target-versus-competitor ordering across representations.

\hypertarget{bounded-within-qwen-comparison}{%
\subsubsection{Bounded Within-Qwen Comparison}\label{bounded-within-qwen-comparison}}

The next question is whether the aggregate ordering across the three configured Qwen levels persisted across the evaluator. Table~\ref{tab:C4} uses initial adjusted mutual information (AMI) because each estimate compares recovered communities with intended attribute assignments for the same 80-item source population. The unit is a generation task under one structural method and embedding configuration. Each level used all 400 planned and observed tasks, and each task-paired contrast used 400 pair-complete tasks across the 20 generator-by-generation-package-by-trait cells, with no missing Qwen evidence. Brackets contain equal-tailed 95\% intervals from 5,000 shared task-bootstrap draws. The comparison describes three complete configurations; it does not isolate parameter count, vector dimension, or any other component cause.

\Needspace{12\baselineskip}
\begingroup
\setlength{\tabcolsep}{1.50pt}
\renewcommand{\arraystretch}{1.05}
\begin{longtable}[]{@{}
  >{\raggedright\arraybackslash}p{(\columnwidth - 12\tabcolsep) * \real{0.1900}}
  >{\raggedleft\arraybackslash}p{(\columnwidth - 12\tabcolsep) * \real{0.1350}}
  >{\raggedleft\arraybackslash}p{(\columnwidth - 12\tabcolsep) * \real{0.1350}}
  >{\raggedleft\arraybackslash}p{(\columnwidth - 12\tabcolsep) * \real{0.1350}}
  >{\raggedleft\arraybackslash}p{(\columnwidth - 12\tabcolsep) * \real{0.1350}}
  >{\raggedleft\arraybackslash}p{(\columnwidth - 12\tabcolsep) * \real{0.1350}}
  >{\raggedleft\arraybackslash}p{(\columnwidth - 12\tabcolsep) * \real{0.1350}}@{}}
\caption{Initial community AMI across the configured Qwen levels.}\label{tab:C4} \\
\toprule\noalign{}
\begin{minipage}[b]{\linewidth}\raggedright
Structural method
\end{minipage} & \begin{minipage}[b]{\linewidth}\raggedleft
Qwen 0.6B
\end{minipage} & \begin{minipage}[b]{\linewidth}\raggedleft
Qwen 4B
\end{minipage} & \begin{minipage}[b]{\linewidth}\raggedleft
Qwen 8B
\end{minipage} & \begin{minipage}[b]{\linewidth}\raggedleft
Qwen 4B - 0.6B
\end{minipage} & \begin{minipage}[b]{\linewidth}\raggedleft
Qwen 8B - 4B
\end{minipage} & \begin{minipage}[b]{\linewidth}\raggedleft
Qwen 8B - 0.6B
\end{minipage} \\
\midrule\noalign{}
\endfirsthead
\toprule\noalign{}
\begin{minipage}[b]{\linewidth}\raggedright
Structural method
\end{minipage} & \begin{minipage}[b]{\linewidth}\raggedleft
Qwen 0.6B
\end{minipage} & \begin{minipage}[b]{\linewidth}\raggedleft
Qwen 4B
\end{minipage} & \begin{minipage}[b]{\linewidth}\raggedleft
Qwen 8B
\end{minipage} & \begin{minipage}[b]{\linewidth}\raggedleft
Qwen 4B - 0.6B
\end{minipage} & \begin{minipage}[b]{\linewidth}\raggedleft
Qwen 8B - 4B
\end{minipage} & \begin{minipage}[b]{\linewidth}\raggedleft
Qwen 8B - 0.6B
\end{minipage} \\
\midrule\noalign{}
\endhead
\bottomrule\noalign{}
\endlastfoot
TMFG & .653 & .691 & .715 & .038 {[}.027, .049{]} & .024 {[}.014, .035{]} & .062 {[}.051, .073{]} \\
Graphical lasso & .758 & .782 & .791 & .024 {[}.015, .033{]} & .009 {[}.001, .016{]} & .033 {[}.024, .042{]} \\
\end{longtable}
\endgroup

The initial aggregate ordering did not persist at every stage or within every content stratum. Final AMI within the Qwen comparison was highest at Qwen 4B under both methods: .906 compared with .899 for Qwen 8B under TMFG, and .891 compared with .886 under graphical lasso. Only 59 of 100 construct-evidence cells were monotonic across the three configured levels. Larger configured models therefore did not produce a uniform larger-is-better result throughout the evaluator.

\hypertarget{broad-geometry-and-local-evaluator-outcomes}{%
\subsection{Broad Geometry and Local Evaluator Outcomes}\label{broad-geometry-and-local-evaluator-outcomes}}

The scientific question here is whether embeddings that organized the complete 80-item populations similarly also preserved the same items and initial community partitions. Geometry agreement uses one generation task's ordering of all 3,160 unique item-pair similarities as its unit; retained-set Jaccard and initial community AMI use the same task under a specified structural method. Across all ten embedding pairs, mean geometry agreement ranged from .733 to .874, while retained-set Jaccard ranged from .410 to .565 under TMFG and from .666 to .755 under graphical lasso. Initial-community correspondence also differed by embedding pair. These outcomes describe different properties: broad pairwise ordering, exact retained identity, and partition correspondence. Higher agreement on one does not establish equivalent downstream decisions or the correctness of either result.

Table~\ref{tab:C5} reports the exact retained-set comparison for every embedding pair under both structural methods. Each row summarizes task-paired source-item sets from the planned 400 generation tasks, with the single BGE-M3/TMFG noncompletion retained as missing where it prevents a pair-complete comparison. Brackets contain equal-tailed 95\% intervals from the existing shared task-bootstrap draws. Jaccard measures exact retained identity and does not indicate which retained set is preferable. Raw retained-set Jaccard is sensitive to the sizes of the compared sets. These comparisons therefore describe the combined consequences of retention volume and exact retained identity rather than agreement after conditioning on retained count. We retain the raw measure because both are consequences of the configured evaluator.

\Needspace{10\baselineskip}
\begin{longtable}[]{@{}
  >{\raggedright\arraybackslash}p{(\columnwidth - 4\tabcolsep) * \real{0.3333}}
  >{\raggedleft\arraybackslash}p{(\columnwidth - 4\tabcolsep) * \real{0.3333}}
  >{\raggedleft\arraybackslash}p{(\columnwidth - 4\tabcolsep) * \real{0.3333}}@{}}
\caption{Raw retained-set overlap across all embedding pairs.}\label{tab:C5} \\
\toprule\noalign{}
\begin{minipage}[b]{\linewidth}\raggedright
Embedding pair
\end{minipage} & \begin{minipage}[b]{\linewidth}\raggedleft
TMFG Jaccard {[}95\% interval{]}
\end{minipage} & \begin{minipage}[b]{\linewidth}\raggedleft
Graphical-lasso Jaccard {[}95\% interval{]}
\end{minipage} \\
\midrule\noalign{}
\endfirsthead
\toprule\noalign{}
\begin{minipage}[b]{\linewidth}\raggedright
Embedding pair
\end{minipage} & \begin{minipage}[b]{\linewidth}\raggedleft
TMFG Jaccard {[}95\% interval{]}
\end{minipage} & \begin{minipage}[b]{\linewidth}\raggedleft
Graphical-lasso Jaccard {[}95\% interval{]}
\end{minipage} \\
\midrule\noalign{}
\endhead
\bottomrule\noalign{}
\endlastfoot
EmbeddingGemma-BGE-M3 & .416 {[}.400, .431{]} & .689 {[}.677, .700{]} \\
Qwen 0.6B-BGE-M3 & .410 {[}.397, .424{]} & .673 {[}.662, .684{]} \\
Qwen 0.6B-EmbeddingGemma & .465 {[}.451, .479{]} & .711 {[}.702, .721{]} \\
Qwen 0.6B-Qwen 4B & .502 {[}.488, .515{]} & .715 {[}.706, .724{]} \\
Qwen 0.6B-Qwen 8B & .499 {[}.486, .511{]} & .708 {[}.700, .717{]} \\
Qwen 4B-BGE-M3 & .424 {[}.410, .437{]} & .676 {[}.665, .686{]} \\
Qwen 4B-EmbeddingGemma & .491 {[}.477, .505{]} & .715 {[}.706, .725{]} \\
Qwen 4B-Qwen 8B & .565 {[}.552, .578{]} & .755 {[}.746, .764{]} \\
Qwen 8B-BGE-M3 & .420 {[}.407, .433{]} & .666 {[}.656, .676{]} \\
Qwen 8B-EmbeddingGemma & .490 {[}.477, .504{]} & .707 {[}.698, .716{]} \\
\end{longtable}

The Qwen 4B-Qwen 8B comparison supplies the clearest already consolidated example. Their geometry agreement was .874 {[}.872, .875{]}, but under TMFG their retained-set Jaccard was .565 {[}.552, .578{]} and their initial-community AMI was .720 {[}.712, .729{]}. The fixed item population could therefore retain a broadly similar semantic ordering while local network relations, recovered boundaries, and threshold-based removal produced different review content.

\hypertarget{selected-task-level-comparisons}{%
\subsubsection{Selected Task-Level Comparisons}\label{selected-task-level-comparisons}}

Table~\ref{tab:C6} asks which paired task-level contrasts anchor the main narrative and whether the representation effect differed across realized source populations. The first four rows use 400 planned, observed, and pair-complete generation tasks with no missing evidence. The generator and generation-package contrasts use 400 planned and observed tasks and 200 pair-complete tasks across two independent 200-task source populations. All intervals use 5,000 shared task-bootstrap draws. The unit is the generation task or paired source population named in the row. These contrasts localize dependence within the configured design; they do not rank generator quality, isolate a prompt-only effect, or establish item validity.

\Needspace{10\baselineskip}
\begingroup
\setlength{\tabcolsep}{2.00pt}
\renewcommand{\arraystretch}{1.05}
\begin{longtable}[]{@{}
  >{\raggedright\arraybackslash}p{(\columnwidth - 2\tabcolsep) * \real{0.8000}}
  >{\raggedleft\arraybackslash}p{(\columnwidth - 2\tabcolsep) * \real{0.2000}}@{}}
\caption{Selected task-level contrasts linking representation, retention, and source-population dependence.}\label{tab:C6} \\
\toprule\noalign{}
\begin{minipage}[b]{\linewidth}\raggedright
Comparison
\end{minipage} & \begin{minipage}[b]{\linewidth}\raggedleft
Estimate {[}95\% interval{]}
\end{minipage} \\
\midrule\noalign{}
\endfirsthead
\toprule\noalign{}
\begin{minipage}[b]{\linewidth}\raggedright
Comparison
\end{minipage} & \begin{minipage}[b]{\linewidth}\raggedleft
Estimate {[}95\% interval{]}
\end{minipage} \\
\midrule\noalign{}
\endhead
\bottomrule\noalign{}
\endlastfoot
Qwen 8B minus Qwen 4B behavioral-indicator scaled target advantage & .009 {[}.005, .013{]} \\
Qwen 4B-Qwen 8B geometry agreement & .874 {[}.872, .875{]} \\
Qwen 4B-Qwen 8B TMFG retained-set Jaccard & .565 {[}.552, .578{]} \\
Qwen 4B-Qwen 8B TMFG initial community AMI & .720 {[}.712, .729{]} \\
Difference between Qwen-generated and Gemma-generated source populations in Qwen 4B-Qwen 8B TMFG retained-set Jaccard & .080 {[}.054, .106{]} \\
Difference between Qwen-generated and Gemma-generated source populations in Qwen 4B-Qwen 8B TMFG initial community AMI & .065 {[}.048, .083{]} \\
Construct-indirect minus AI-GENIE source Qwen 8B behavioral-indicator scaled target advantage & .222 {[}.202, .242{]} \\
\end{longtable}
\endgroup

The last three contrasts show why generation and later evaluation should remain distinct but linked. The source population moderated how consistently two representations preserved items and communities, while the compound generation packages changed the kind of anchor alignment visible in the generated content. Because each package changed instructions and candidate gating together, the package contrast cannot be reduced to label prohibition or behavioral guidance alone.

\hypertarget{what-structural-reduction-preserved-and-lost}{%
\subsection{What Structural Reduction Preserved and Lost}\label{what-structural-reduction-preserved-and-lost}}

The final question is how a structurally simpler retained population differed from its 80-item source population. Table~\ref{tab:C7} uses one structural evaluation as its unit and gives equal weight to the fixed generation design. Each embedding-method row represents 400 planned tasks except that BGE-M3 under TMFG excludes one repeatable BootEGA noncompletion from its evaluable evidence. Initial communities and AMI use all 80 source items; final communities, final AMI, and final item count use each configuration's separately retained population. Change is final minus initial. The table describes a reduction pathway across changing item populations and cannot be interpreted as a validity gain or a direct like-for-like reclustering effect.

\Needspace{10\baselineskip}
\begingroup
\setlength{\tabcolsep}{1.50pt}
\renewcommand{\arraystretch}{1.05}
\begin{longtable}[]{@{}
  >{\raggedright\arraybackslash}p{(\columnwidth - 14\tabcolsep) * \real{0.1700}}
  >{\raggedright\arraybackslash}p{(\columnwidth - 14\tabcolsep) * \real{0.1700}}
  >{\raggedleft\arraybackslash}p{(\columnwidth - 14\tabcolsep) * \real{0.1000}}
  >{\raggedleft\arraybackslash}p{(\columnwidth - 14\tabcolsep) * \real{0.1000}}
  >{\raggedleft\arraybackslash}p{(\columnwidth - 14\tabcolsep) * \real{0.1000}}
  >{\raggedleft\arraybackslash}p{(\columnwidth - 14\tabcolsep) * \real{0.1000}}
  >{\raggedleft\arraybackslash}p{(\columnwidth - 14\tabcolsep) * \real{0.1200}}
  >{\raggedleft\arraybackslash}p{(\columnwidth - 14\tabcolsep) * \real{0.1400}}@{}}
\caption{Intended-attribute community correspondence and retained population before and after reduction.}\label{tab:C7} \\
\toprule\noalign{}
\multicolumn{2}{c}{} & \multicolumn{2}{c}{Communities} & \multicolumn{3}{c}{AMI} & \multicolumn{1}{c}{} \\
\cmidrule(lr){3-4}\cmidrule(lr){5-7}
\begin{minipage}[b]{\linewidth}\raggedright
Structural method
\end{minipage} & \begin{minipage}[b]{\linewidth}\raggedright
Embedding
\end{minipage} & \begin{minipage}[b]{\linewidth}\centering
Initial
\end{minipage} & \begin{minipage}[b]{\linewidth}\centering
Final
\end{minipage} & \begin{minipage}[b]{\linewidth}\centering
Initial
\end{minipage} & \begin{minipage}[b]{\linewidth}\centering
Final
\end{minipage} & \begin{minipage}[b]{\linewidth}\centering
Change
\end{minipage} & \begin{minipage}[b]{\linewidth}\centering
Final items
\end{minipage} \\
\midrule\noalign{}
\endfirsthead
\toprule\noalign{}
\multicolumn{2}{c}{} & \multicolumn{2}{c}{Communities} & \multicolumn{3}{c}{AMI} & \multicolumn{1}{c}{} \\
\cmidrule(lr){3-4}\cmidrule(lr){5-7}
\begin{minipage}[b]{\linewidth}\raggedright
Structural method
\end{minipage} & \begin{minipage}[b]{\linewidth}\raggedright
Embedding
\end{minipage} & \begin{minipage}[b]{\linewidth}\centering
Initial
\end{minipage} & \begin{minipage}[b]{\linewidth}\centering
Final
\end{minipage} & \begin{minipage}[b]{\linewidth}\centering
Initial
\end{minipage} & \begin{minipage}[b]{\linewidth}\centering
Final
\end{minipage} & \begin{minipage}[b]{\linewidth}\centering
Change
\end{minipage} & \begin{minipage}[b]{\linewidth}\centering
Final items
\end{minipage} \\
\midrule\noalign{}
\endhead
\bottomrule\noalign{}
\endlastfoot
TMFG & BGE-M3 & 5.636 & 3.756 & .581 & .870 & .289 & 38.132 \\
TMFG & EmbeddingGemma & 5.385 & 3.930 & .662 & .902 & .241 & 45.325 \\
TMFG & Qwen 0.6B & 5.408 & 3.885 & .653 & .896 & .244 & 44.045 \\
TMFG & Qwen 4B & 5.360 & 4.032 & .691 & .906 & .215 & 49.140 \\
TMFG & Qwen 8B & 5.327 & 4.060 & .715 & .899 & .184 & 51.133 \\
\mbox{Graphical lasso} & BGE-M3 & 4.827 & 4.135 & .692 & .835 & .143 & 59.990 \\
\mbox{Graphical lasso} & EmbeddingGemma & 4.875 & 4.220 & .767 & .893 & .126 & 63.592 \\
\mbox{Graphical lasso} & Qwen 0.6B & 4.888 & 4.202 & .758 & .876 & .118 & 62.013 \\
\mbox{Graphical lasso} & Qwen 4B & 4.928 & 4.185 & .782 & .891 & .109 & 63.605 \\
\mbox{Graphical lasso} & Qwen 8B & 5.060 & 4.192 & .791 & .886 & .095 & 63.285 \\
\end{longtable}
\endgroup

Reduction usually increased correspondence with the intended attribute assignments, but that change was purchased by changing the item population. AMI decreased in 59 of 1,999 evaluable TMFG contexts and 217 of 2,000 graphical-lasso contexts. Initial and final community counts also moved toward four: exact four-community solutions rose from 32.93\% initially to 73.02\% finally. Four communities did not guarantee recovery of the intended organization. Under TMFG, Qwen 4B produced four final communities in 311 tasks, but only 106 matched the intended attributes perfectly; six of those exact-four contexts had already lost every item assigned to one intended attribute. Qwen 8B produced four final communities in 304 tasks, 114 of which matched perfectly; five exact-four contexts had lost an intended attribute completely.

Content coverage and exact-item identity remained separate from community correspondence. Every task began with 20 items for each of four attributes. Under TMFG, BGE-M3 ended with 119 empty attribute cells among 1,596 evaluable cells, compared with 27 of 1,600 under Qwen 8B. Graphical-lasso zero coverage ranged from 0.2\% under Qwen 4B to 1.3\% under BGE-M3. Even where an attribute remained represented, cross-embedding retained-set Jaccard showed that a similar amount of retained content could consist of different statements.

The companion partition metrics set a further boundary on the AMI narrative. At the aggregate configuration level, AMI, adjusted Rand index (ARI), and normalized mutual information (NMI) agreed on all 45 pairwise orderings for change, while differing on one of 45 initial orderings and two of 45 final orderings. At the individual-task level, after absolute changes below 1e-12 were treated as numerical zero, the three metrics assigned the same sign to change in 3,882 of 3,999 evaluable structural evaluations and different signs in 117; all 29 reclassified cases had identical released ARI endpoints of 1.0 and change values at floating-point precision. AMI can therefore lead the summary without implying that metric choice is irrelevant in every realized task.

The single structural noncompletion remains part of this evidence boundary. It occurred for one Qwen-generated, AI-GENIE-source, Agreeableness task under BGE-M3 and TMFG. Its partial events did not establish retention or nonretention, so the context and its 80 item outcomes remain unknown wherever complete structural evidence is required. The missing result is one analytic noncompletion, not 80 rejected items.

\clearpage
\hypertarget{appendix-d.-study-2-supporting-evidence}{%
\section{Study 2 Supporting Evidence}\label{appendix-d.-study-2-supporting-evidence}}

This appendix provides the configuration-level and generation-task resampling evidence needed to distinguish form completeness from exact wording stability. It is limited to the two declared eligibility policies applied to the same generated items, embeddings, structural results, anchor ranking, assignment procedure, and form quotas. The evidence describes what the configured evaluator placed before psychometricians; it does not establish that either policy produced higher-quality or psychometrically valid items.

\hypertarget{configuration-level-policy-and-resampling-evidence}{%
\subsection{Configuration-Level Policy and Resampling Evidence}\label{configuration-level-policy-and-resampling-evidence}}

The scientific question is how much requiring support from both structural methods contracted the form-eligible item pool, whether that contraction prevented complete candidate forms, and how much exact primary wording remained shared within each configuration. Table~\ref{tab:D1} uses one generator-by-generation-package-by-embedding configuration as its full-data unit. Each configuration contains 100 generation tasks and 8,000 selected item occurrences and seeks 40 primary and 40 alternate positions across 20 content cells. Eligibility reduction uses the inclusive-eligible occurrences with complete paired method evidence within that configuration as its denominator. Shared primary and full-data Jaccard compare normalized wording in the two 40-statement primary forms. Resampling values summarize 1,000 paired generation-task draws per configuration, except for the conditionally evaluable BGE-M3 case identified below. These quantities show policy dependence within the configured evaluator, not policy optimality, participant-sampling uncertainty, or item validity.

\Needspace{14\baselineskip}
\begingroup
\setlength{\tabcolsep}{1.50pt}
\renewcommand{\arraystretch}{1.05}
\begin{longtable}[]{@{}
  >{\raggedright\arraybackslash}p{(\columnwidth - 14\tabcolsep) * \real{0.0800}}
  >{\raggedright\arraybackslash}p{(\columnwidth - 14\tabcolsep) * \real{0.0900}}
  >{\raggedright\arraybackslash}p{(\columnwidth - 14\tabcolsep) * \real{0.1400}}
  >{\raggedleft\arraybackslash}p{(\columnwidth - 14\tabcolsep) * \real{0.1200}}
  >{\raggedleft\arraybackslash}p{(\columnwidth - 14\tabcolsep) * \real{0.1200}}
  >{\raggedleft\arraybackslash}p{(\columnwidth - 14\tabcolsep) * \real{0.1200}}
  >{\raggedleft\arraybackslash}p{(\columnwidth - 14\tabcolsep) * \real{0.1000}}
  >{\raggedright\arraybackslash}p{(\columnwidth - 14\tabcolsep) * \real{0.2300}}@{}}
\caption{Eligibility contraction and exact primary-wording overlap across 20 form configurations. Source and Indirect denote the complete AI-GENIE-source and construct-indirect prompt-and-gate packages. NE indicates that complete structural evidence was unavailable. Brackets contain the middle 95\% of evaluable generation-task resampling draws.}\label{tab:D1} \\
\toprule\noalign{}
\begin{minipage}[b]{\linewidth}\raggedright
Generator
\end{minipage} & \begin{minipage}[b]{\linewidth}\raggedright
Package
\end{minipage} & \begin{minipage}[b]{\linewidth}\raggedright
Embedding
\end{minipage} & \begin{minipage}[b]{\linewidth}\raggedleft
Eligibility reduction
\end{minipage} & \begin{minipage}[b]{\linewidth}\raggedleft
Shared primary / 40
\end{minipage} & \begin{minipage}[b]{\linewidth}\raggedleft
Policy-specific per policy
\end{minipage} & \begin{minipage}[b]{\linewidth}\raggedleft
Full-data Jaccard
\end{minipage} & \begin{minipage}[b]{\linewidth}\raggedright
Resampled median {[}middle 95\%{]}
\end{minipage} \\
\midrule\noalign{}
\endfirsthead
\toprule\noalign{}
\begin{minipage}[b]{\linewidth}\raggedright
Generator
\end{minipage} & \begin{minipage}[b]{\linewidth}\raggedright
Package
\end{minipage} & \begin{minipage}[b]{\linewidth}\raggedright
Embedding
\end{minipage} & \begin{minipage}[b]{\linewidth}\raggedleft
Eligibility reduction
\end{minipage} & \begin{minipage}[b]{\linewidth}\raggedleft
Shared primary / 40
\end{minipage} & \begin{minipage}[b]{\linewidth}\raggedleft
Policy-specific per policy
\end{minipage} & \begin{minipage}[b]{\linewidth}\raggedleft
Full-data Jaccard
\end{minipage} & \begin{minipage}[b]{\linewidth}\raggedright
Resampled median {[}middle 95\%{]}
\end{minipage} \\
\midrule\noalign{}
\endhead
\bottomrule\noalign{}
\endlastfoot
Qwen & Source & BGE-M3 & NE & NE & NE & NE & .404 {[}.310, .538{]} (359 draws) \\
Qwen & Source & Qwen 0.6B & 35.9\% & 22 & 18 & .379 & .481 {[}.379, .601{]} \\
Qwen & Source & Qwen 4B & 26.6\% & 30 & 10 & .600 & .600 {[}.481, .702{]} \\
Qwen & Source & Qwen 8B & 22.0\% & 33 & 7 & .702 & .702 {[}.633, .818{]} \\
Qwen & Source & EmbeddingGemma & 35.8\% & 24 & 16 & .429 & .481 {[}.379, .600{]} \\
Qwen & Indirect & BGE-M3 & 39.5\% & 26 & 14 & .481 & .481 {[}.356, .600{]} \\
Qwen & Indirect & Qwen 0.6B & 31.5\% & 29 & 11 & .569 & .538 {[}.429, .667{]} \\
Qwen & Indirect & Qwen 4B & 26.1\% & 31 & 9 & .633 & .667 {[}.538, .778{]} \\
Qwen & Indirect & Qwen 8B & 21.4\% & 35 & 5 & .778 & .778 {[}.667, .860{]} \\
Qwen & Indirect & EmbeddingGemma & 34.6\% & 32 & 8 & .667 & .667 {[}.538, .778{]} \\
Gemma & Source & BGE-M3 & 53.0\% & 23 & 17 & .404 & .379 {[}.290, .509{]} \\
Gemma & Source & Qwen 0.6B & 41.3\% & 30 & 10 & .600 & .600 {[}.481, .702{]} \\
Gemma & Source & Qwen 4B & 35.3\% & 30 & 10 & .600 & .538 {[}.429, .667{]} \\
Gemma & Source & Qwen 8B & 30.5\% & 29 & 11 & .569 & .600 {[}.509, .739{]} \\
Gemma & Source & EmbeddingGemma & 41.9\% & 25 & 15 & .455 & .455 {[}.356, .569{]} \\
Gemma & Indirect & BGE-M3 & 47.2\% & 26 & 14 & .481 & .455 {[}.356, .569{]} \\
Gemma & Indirect & Qwen 0.6B & 39.5\% & 27 & 13 & .509 & .538 {[}.455, .633{]} \\
Gemma & Indirect & Qwen 4B & 28.3\% & 30 & 10 & .600 & .633 {[}.538, .778{]} \\
Gemma & Indirect & Qwen 8B & 27.0\% & 31 & 9 & .633 & .633 {[}.509, .739{]} \\
Gemma & Indirect & EmbeddingGemma & 38.9\% & 26 & 14 & .481 & .509 {[}.404, .667{]} \\
\end{longtable}
\endgroup

Across the 19 configurations with complete evidence for full form construction, agreement removed 42,830 of the 124,408 item occurrences that were form-eligible under the inclusive policy, yielding a pooled item-occurrence reduction of 34.4\%; configuration-specific reductions ranged from 21.4\% to 53.0\%. Both policies nevertheless filled every primary and alternate position in every evaluable configuration. Complete content-cell coverage therefore coexisted with different exact wording: the two primary forms shared a median of 29 of 40 statements, leaving a median of 11 statements specific to each policy, and no evaluable agreement form reproduced the inclusive primary set exactly. This feasibility result is conditional on the large observed candidate pools and should not be generalized to smaller or more restricted pools.

\hypertarget{what-persisted-across-generation-task-mixtures}{%
\subsection{What Persisted Across Generation-Task Mixtures}\label{what-persisted-across-generation-task-mixtures}}

The resampling evidence asks whether the full-data policy relationship depended on the particular 20 generation tasks realized for each generator, package, and trait. Of 20,000 planned paired configuration draws, 19,359 were evaluable, and every evaluable draw filled all 40 primary and 40 alternate positions under both policies. Configuration-specific median primary-wording Jaccard ranged from .379 to .778. These are repeated generation-task mixtures, not respondent samples or confidence intervals for population psychometrics.

The Qwen/AI-GENIE-source/BGE-M3 configuration remained non-evaluable in the full data because one of its 100 source tasks contained the single missing TMFG result. Of its 1,000 resamples, 359 did not draw that task and were evaluable, while 641 drew it at least once and remained non-evaluable. The conditional 359-draw distribution describes only complete-evidence task mixtures; it does not recover the missing structural result or supply a full-data baseline.

Representation changes produced greater exact-wording divergence than the post-method policy contrast. Within the same source population and eligibility policy, 36 of 40 planned inclusive-form embedding pairs were evaluable and shared a median of 6 of 40 primary statements, with a median Jaccard of .081. Within the same embedding configuration, the 19 evaluable inclusive-versus-agreement comparisons shared a median of 29 of 40 primary statements and had a median Jaccard of .569. The units are matched pairs of 40-statement primary forms. This comparison describes the bundled representation-plus-structural configuration against a narrower post-method policy change; it does not isolate a causal embedding-model component.

The secondary diagnostics did not supply a consistent compensating advantage for agreement. Inclusive primary forms averaged 16.53 warning-bearing items and agreement forms averaged 15.79, but the paired direction varied across configurations. Primary forms drew from approximately 34 of the 100 source tasks under either policy, and no single task supplied more than three of 40 primary statements. Selected items had survived UVA under both methods, exact normalized-wording duplicates were excluded by construction, and source concentration remained similar. The demonstrated policy consequence was therefore contraction of eligibility and different expert-facing wording, not a consistent improvement in warning burden, local-dependence evidence, or source diversity.

\clearpage
\hypertarget{appendix-e.-source-pool-depth-sensitivity-analysis}{%
\section{Source-Pool Depth Sensitivity Analysis}\label{appendix-e.-source-pool-depth-sensitivity-analysis}}

Source-population depth determines both which candidate statements are available for reduction and the population against which each statement is evaluated. In a network-integrated evaluator, adding items can therefore do more than provide additional options: it can change redundancy decisions, community assignments, retention, and the ranking of candidates that were already present. We treated depth as a measurement-design choice and examined it before fixing the 80-item source populations used in the primary studies.

The calibration asked whether larger source populations continued to contribute distinct candidate content, whether the treatment of already available candidates became more stable, and whether later additions became increasingly similar to earlier content. The decision was based on the joint pattern across these questions rather than a single cutoff or overlap statistic.

\hypertarget{exact-nested-calibration-design}{%
\subsection{Exact Nested Calibration Design}\label{exact-nested-calibration-design}}

The bounded calibration included Openness and Neuroticism and crossed the same two generators and two generation packages used in the primary studies. Each generator-by-package-by-trait cell contained 12 independently seeded generation tasks, yielding 96 tasks. A task produced an ordered population of 100 gate-eligible items, balanced as 25 items for each of the trait's four attributes. Within each attribute, the first 10, 15, 20, and 25 items to fill the eligibility quota defined exact nested source populations of 40, 60, 80, and 100 items. This order records when an item became eligible; it is not an estimated quality rank.

The complete 100-item population from each task was embedded once in each of the five embedding spaces used in the paper. The smaller populations were exact item-column slices of those same matrices, so adjacent comparisons held item wording and vector values fixed while changing the candidate population supplied to the structural evaluator. Each depth was evaluated separately with TMFG and graphical lasso, producing 960 planned generation-task-by-embedding-by-method contexts per depth. The configured UVA, EGA, bootstrap EGA, and retention settings were held constant.

Adjacent comparisons included a context only when both depths produced complete final-item evidence for the same task, embedding, and structural method. This yielded 950 paired contexts for 40-60, 959 for 60-80, and 958 for 80-100. Graphical lasso completed all 480 contexts at every depth. TMFG had 9 analytic noncompletions at depth 40, 1 at depth 60, none at depth 80, and 2 at depth 100; partial outputs were not treated as complete or imputed.

The calibration's 12 generation tasks per design cell should not be confused with the 20 independently seeded repetitions used in the primary studies. This appendix evaluates source-population depth. It does not evaluate the choice of 20 repetitions.

\hypertarget{evidence-across-adjacent-depths}{%
\subsection{Evidence Across Adjacent Depths}\label{evidence-across-adjacent-depths}}

Table~\ref{tab:E1} separates three kinds of sensitivity. Retained-item Jaccard asks whether the same items survived at adjacent depths. Adjusted Rand index (ARI) asks whether items retained at both depths occupied similar communities after accounting for label permutation. The diagnostic review slate asks whether a deterministic two-item-per-cell projection selected the same normalized statements after the population changed. These slates pooled evidence across the 12 generation tasks and omitted the construct-anchor evidence used in Study 1; they are sensitivity diagnostics, not Study 2 candidate forms.

\Needspace{12\baselineskip}
\begingroup
\setlength{\tabcolsep}{1.25pt}
\renewcommand{\arraystretch}{1.00}
\begin{longtable}[]{@{}
  >{\raggedright\arraybackslash}p{(\columnwidth - 6\tabcolsep) * \real{0.4600}}
  >{\raggedleft\arraybackslash}p{(\columnwidth - 6\tabcolsep) * \real{0.1800}}
  >{\raggedleft\arraybackslash}p{(\columnwidth - 6\tabcolsep) * \real{0.1800}}
  >{\raggedleft\arraybackslash}p{(\columnwidth - 6\tabcolsep) * \real{0.1800}}@{}}
\caption{Retention, shared-item community, and diagnostic-slate sensitivity across adjacent source-population depths.}\label{tab:E1} \\
\toprule\noalign{}
\begin{minipage}[b]{\linewidth}\raggedright
Outcome
\end{minipage} & \begin{minipage}[b]{\linewidth}\raggedleft
40-60
\end{minipage} & \begin{minipage}[b]{\linewidth}\raggedleft
60-80
\end{minipage} & \begin{minipage}[b]{\linewidth}\raggedleft
80-100
\end{minipage} \\
\midrule\noalign{}
\endfirsthead
\toprule\noalign{}
\begin{minipage}[b]{\linewidth}\raggedright
Outcome
\end{minipage} & \begin{minipage}[b]{\linewidth}\raggedleft
40-60
\end{minipage} & \begin{minipage}[b]{\linewidth}\raggedleft
60-80
\end{minipage} & \begin{minipage}[b]{\linewidth}\raggedleft
80-100
\end{minipage} \\
\midrule\noalign{}
\endhead
\bottomrule\noalign{}
\endlastfoot
Complete paired method contexts & 950 & 959 & 958 \\
Retained-item Jaccard, mean & .4484 & .5330 & .5553 \\
Shared-item community ARI, mean (defined \emph{n}) & .9462 (948) & .9679 (959) & .9736 (958) \\
Retained-item count difference, larger minus smaller, mean & 15.42 & 13.67 & 10.86 \\
Diagnostic two-item-slate Jaccard, mean (160 cells) & .3375 & .3354 & .3125 \\
Newly exposed band among larger-only slate entries & 63/187 (33.7\%) & 62/187 (33.2\%) & 37/189 (19.6\%) \\
Shared-rank reselection among larger-only slate entries & 105/187 (56.1\%) & 109/187 (58.3\%) & 129/189 (68.3\%) \\
Maximum prior-band cosine for newly exposed items, median (1,920 occurrences) & .7671 & .7862 & .8072 \\
\end{longtable}
\endgroup

All summaries are descriptive, and contexts sharing a generation task are dependent. Slate-origin counts are candidate-cell occurrences: the same normalized statement selected in different embedding-specific cells contributes once in each cell. Two 40-60 contexts shared no retained items, so their ARI values were undefined.

Retained populations became more similar across successive depth contrasts, and the communities assigned to shared retained items agreed closely throughout. The increase in the mean number of retained items nevertheless declined from 15.42 to 10.86 across the curve. The final step was also method-dependent: mean retained-item Jaccard decreased slightly from .4670 to .4526 for TMFG between the 60-80 and 80-100 contrasts, while it increased from .5989 to .6575 for graphical lasso. Thus, the pooled pattern did not identify a method-independent point at which adding candidates ceased to matter.

The small diagnostic slates remained sensitive at every depth. Their mean overlap did not increase with the broader retained-set overlap, illustrating that stability of a larger retained population does not imply stability at a narrow selection boundary. Candidate provenance clarifies what changed. The bands added in the 40-60 and 60-80 expansions directly supplied approximately one third of the larger-pool-only slate entries. In the 80-100 expansion, direct contributions from the new ranks fell to 19.6\%, while reselection of statements already available at depth 80 rose to 68.3\%. The final added band also had the greatest semantic proximity to earlier same-attribute items. Its median maximum cosine to an earlier item was .8072, compared with .7671 and .7862 for the preceding added bands. These are proximity summaries rather than evidence that statements are interchangeable.

\hypertarget{why-80-items-was-selected}{%
\subsection{Why 80 Items Was Selected}\label{why-80-items-was-selected}}

The combined evidence supported 80 items, balanced as 20 gate-eligible statements per attribute, as a pragmatic source-population depth for the configured studies. Stopping at 60 would have excluded ranks 16-20, a band that contributed new diagnostic-slate content at nearly the same rate as ranks 11-15. Expanding from 80 to 100 continued to alter the evaluator, but the added ranks were more semantically proximal to earlier candidates and supplied a smaller share of the new slate content. Most changes in the final comparison instead arose because the larger population changed the reduction and ranking context around candidates that were already available at depth 80. The additional depth therefore did not stabilize the narrow selection boundary even as it continued to increase the number of retained items.

This was a descriptive, multi-criterion decision. The calibration did not fit a formal elbow estimator, test equivalence between depths, or apply a universal Jaccard, cosine, or community-agreement threshold. It estimates sensitivity across exact eligibility-order prefixes from two traits, two generators, two generation packages, five embedding spaces, and two structural methods. Transferring the selected depth to all five traits and to the primary studies' 20-repetition design is a configured workflow choice rather than a separately estimated generalization. The results support 80 items as a defensible operating point for this paper; they do not establish that 80 items are universally optimal, exhaust a trait's content, or produce respondent-valid scales.

\end{document}